\documentclass[sigconf]{acmart}
\AtBeginDocument{%
  }

\acmConference[KDD '27]{ACM SIGKDD Conference on Knowledge Discovery and Data Mining}{August 1--5, 2027}{San Jose, CA, USA}
\acmYear{2027}
\usepackage{amsmath}
\usepackage{algorithm}
\usepackage{algorithmicx}
\usepackage{algpseudocode}
\usepackage{makecell}
\usepackage{multirow}
\usepackage{colortbl}
\usepackage{soul}
\usepackage{placeins}
\usepackage{pifont}

\newcommand{\hlc}[2][yellow]{{\sethlcolor{#1}\hl{#2}}}

\definecolor{added-color}{RGB}{0,128,0}
\definecolor{removed-color}{RGB}{200,0,0}
\providecommand{\showtrackedchanges}{1}
\newif\ifshowchanges
\ifnum\showtrackedchanges=1
  \showchangestrue
\else
  \showchangesfalse
\fi

\definecolor{jitter-color}{HTML}{4B54E1}
\definecolor{noise-color}{HTML}{93FFF0}
\definecolor{overexposure-color}{HTML}{CDE55D}
\definecolor{blur-color}{HTML}{E45C5C}
\definecolor{lowlight-color}{HTML}{35E344}
\definecolor{srartifact-color}{HTML}{C000A0}

\begin{document}

\title[SR-Ground: Image Quality Grounding for Super-Resolved Content]{SR-Ground: Image Quality Grounding\\for Super-Resolved Content}

\author{Artem Borisov}
\email{artem.borisov@graphics.cs.msu.ru}
\affiliation{%
  \institution{Lomonosov Moscow State University}
  \city{Moscow}
  \country{Russia}
}

\author{Evgeney Bogatyrev}
\email{evgeney.bogatyrev@graphics.cs.msu.ru}
\affiliation{%
  \institution{MSU Institute for AI, Lomonosov Moscow State University}
  \city{Moscow}
  \country{Russia}
}

\author{Khaled Abud}
\email{khaled.abud@graphics.cs.msu.ru}
\affiliation{%
  \institution{MSU Institute for AI, Lomonosov Moscow State University}
  \city{Moscow}
  \country{Russia}
}

%\author{Nikita Kukuzei}
%\email{kukuzeyna@my.msu.ru}
%\affiliation{%
%  \institution{Lomonosov Moscow State University}
%  \city{Moscow}
%  \country{Russia}
%}

\author{Dmitriy Vatolin}
\email{dmitriy@graphics.cs.msu.ru}
\affiliation{%
  \institution{MSU Institute for AI, Lomonosov Moscow State University}
  \city{Moscow}
  \country{Russia}
}

\renewcommand{\shortauthors}{Borisov et al.}

\begin{abstract}
Super-Resolution (SR) has advanced rapidly in recent years, with diffusion-based models achieving unprecedented fidelity at the cost of introducing new types of visual artifacts. While existing Image Quality Assessment (IQA) methods provide holistic quality scores, they lack interpretability and fail to distinguish between different artifact types arising from modern SR approaches.

To address this gap, we introduce SR-Ground, a large-scale dataset specifically designed for fine-grained artifact segmentation in super-resolved images. The dataset comprises images processed by a diverse set of state-of-the-art SR models, with pixel-level annotations for multiple artifact categories. We conduct a large-scale crowdsourcing study involving 1,062 participants to validate and refine automatically generated segmentations, resulting in a high-quality dataset of 63,000 images spanning 6 distinct artifact types.

We demonstrate that training IQA models with grounding capabilities on SR-Ground significantly improves performance on downstream tasks. Furthermore, we introduce a fine-tuning pipeline that leverages our grounding model to reduce perceptible artifacts in SR outputs, showcasing the practical utility of our dataset. On a separate benchmark of 1,000 outputs from five unseen SR methods, an Low-Resolution-referenced grounding model significantly improves prominence alignment over its no-reference counterpart and performs best on native real-world datasets of Low-Resolution images without High-Resolution Ground-Truth.
\end{abstract}




\maketitle

\begin{figure}[h!]
\includegraphics[width=0.9\linewidth]{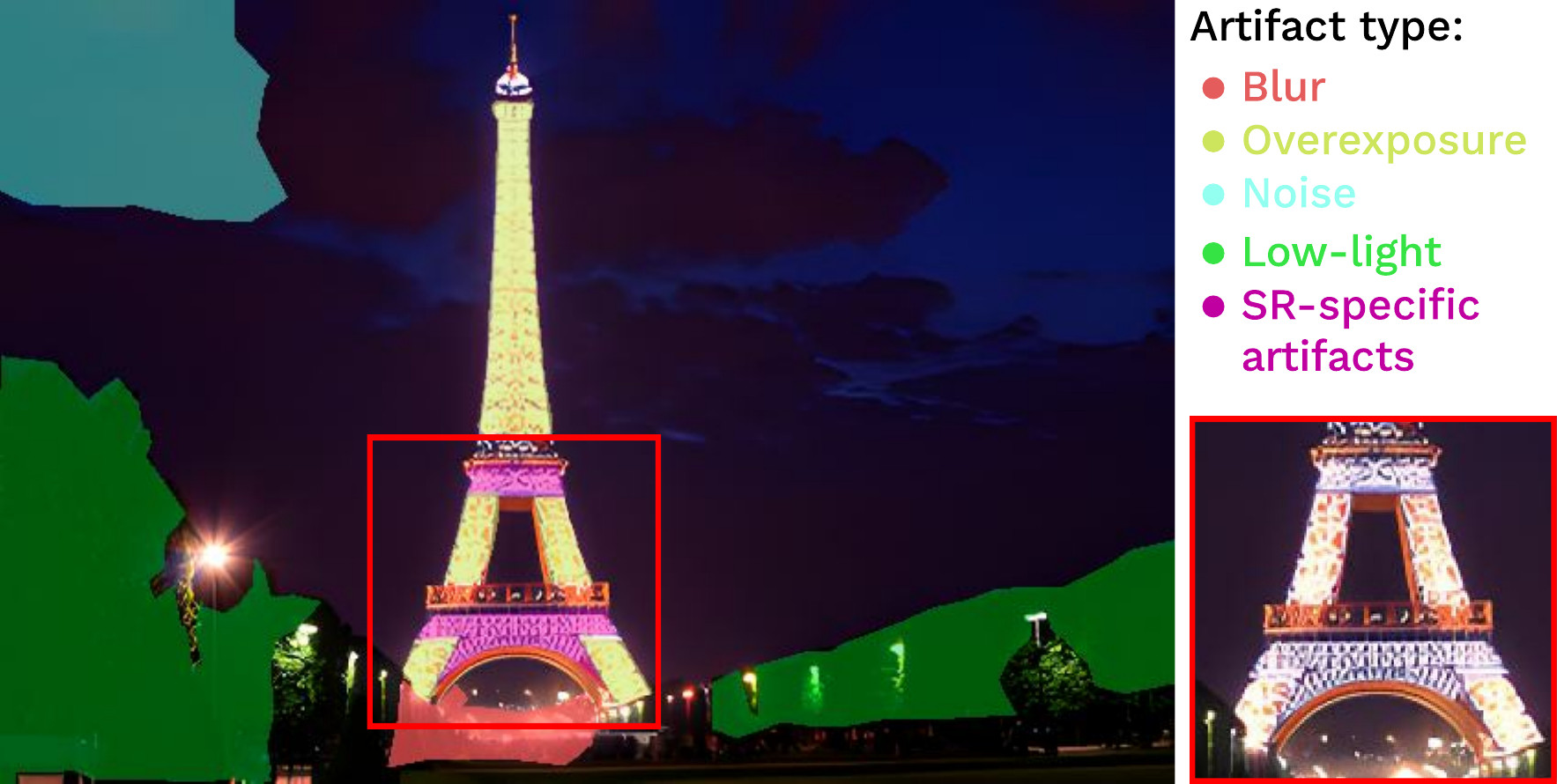}
\caption{An example of Image Quality Grounding. Zoomed part highlights SR-specific artifacts.}
\Description{A super-resolved face crop with a highlighted local distortion, accompanied by an artifact probability map and a binary grounding mask.}
\label{fig:grounding-example}
\end{figure}

\footnotetext{The research was carried out using the MSU-270 supercomputer of Lomonosov Moscow State University.}

\section{Introduction}

Recent advances in Super-Resolution (SR), particularly diffusion-based approaches, have led to substantial improvements in perceptual quality~\cite{yang2023pasd, yu2024scaling, duan2025dit4sr}. However, these models often introduce subtle yet perceptually significant visual artifacts~\cite{Bogatyrev_2024_BMVC}, such as unnatural textures, distorted facial structures, and locally inconsistent patterns. A number of recent works have explored detecting and localizing such artifacts~\cite{desra, liang2022details, molodetskikh2026srprominencecrowdsourcedprotocoldataset}, as well as mitigating them during SR reconstruction~\cite{molodetskikh2026srprominencecrowdsourcedprotocoldataset}.

\begin{figure*}[h!]
\includegraphics[width=0.95\linewidth]{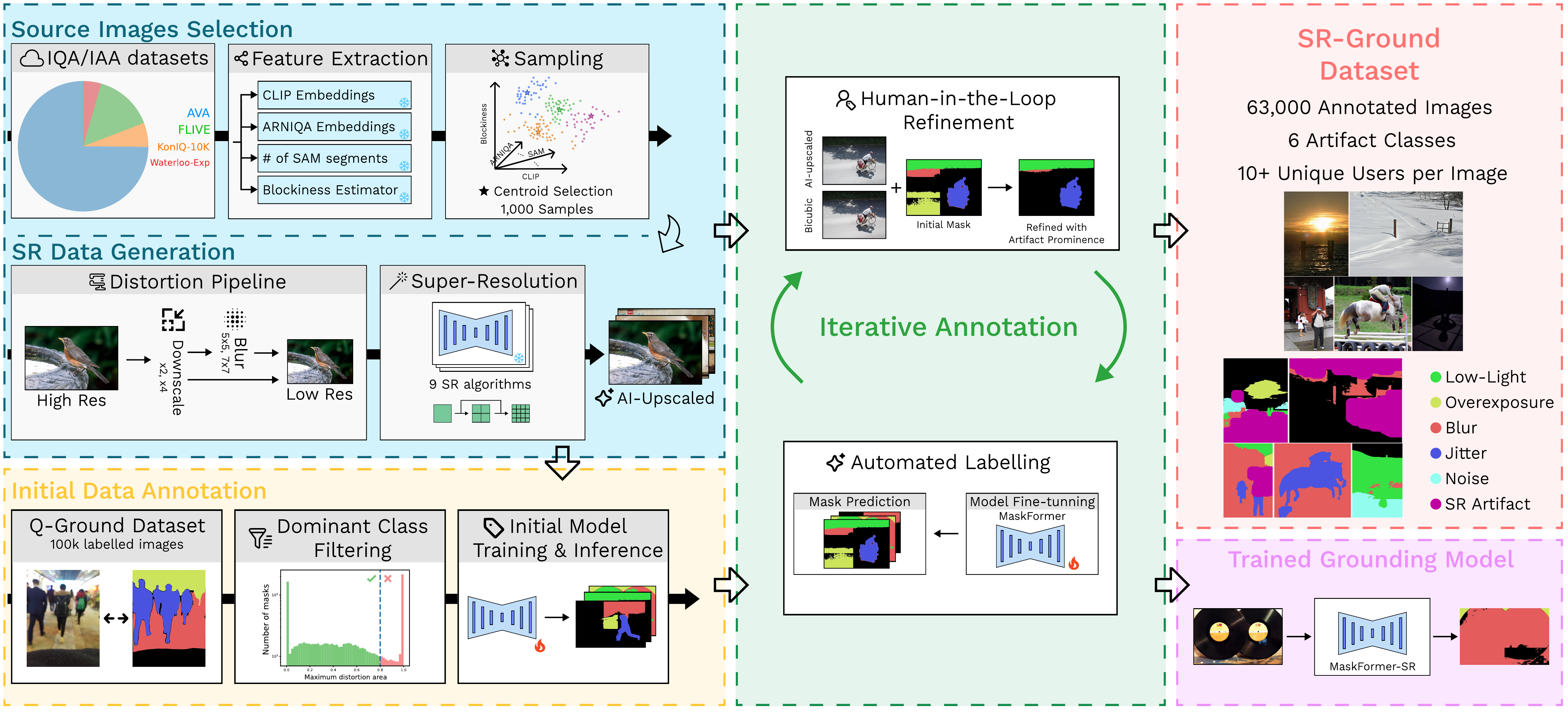}
\caption{SR-Ground Iterative Dataset Curation Pipeline}
\Description{Flowchart of source-image selection, super-resolution data generation, initial automatic annotation, crowdsourced refinement, and iterative model fine-tuning.}
\label{fig:iterative-pipeline}
\end{figure*}

Importantly, SR artifacts differ from distortions commonly observed in natural images or those produced by other generative models. As a result, existing \textbf{Image Quality Assessment} (IQA) metrics often fail to reliably evaluate super-resolved content~\cite{srqabenchmark}. Despite their impact on visual fidelity, these artifacts remain difficult to quantify and analyze using current IQA methods, which typically provide only a single global quality score per image and lack interpretability~\cite{zhang2018perceptual, chen2024topiq, wu2023qalign}. For the remainder of this paper, we refer to such distortions as \emph{SR-specific artifacts}.

Operationally, an SR-specific artifact is a spatially localized distortion introduced by an SR method. Typical cases include unnatural textures, distorted facial structures, and hallucinated details. We retain a single class because the literature does not yet provide a stable pixel-level subtype taxonomy~\cite{desra, molodetskikh2026srprominencecrowdsourcedprotocoldataset}; developing such a taxonomy is outside the scope of this dataset release.

Visual Quality Reasoning models~\cite{wu2023qinstruct, jia2024vqa} provide a detailed description of image distortions; however, they struggle to localize these distortions and therefore are not suitable for dense tasks. At the same time, existing \textbf{Image Quality Grounding} (IQG) and segmentation approaches are not designed to capture SR-specific distortions, limiting their applicability for fine-grained artifact analysis~\cite{chen2024qground, chen2026grounding}. This gap motivates the need for a dataset that targets the localization and categorization of artifacts produced by SR models.

We propose \textbf{SR-Ground}, a large-scale dataset tailored for fine-grained artifact and real distortions grounding in super-resolved images. SR-Ground focuses on distortions introduced by CNN-based, GAN-based, transformer-based and diffusion-based SR models, which exhibit diverse and previously underexplored artifact patterns. The dataset provides pixel-level annotations across multiple artifact categories, enabling detailed spatial analysis. We construct SR-Ground through an iterative data generation and refinement pipeline, combining automated annotation with large-scale crowdsourced validation to ensure high-quality and consistent labels.

Using this dataset, we demonstrate that training segmentation models and adapting them to SR artifacts significantly improves performance in 
the image-quality-grounding task. As a secondary application, we use the trained grounding model as a frozen spatial controller in an OSEDiff fine-tuning pipeline for the task of removing and adding various types of distortions.

Our \textbf{main contributions} are as follows:
\begin{itemize}
    \item We introduce \textbf{SR-Ground}, a large-scale dataset for pixel-level grounding of SR-specific visual artifacts. This dataset contains 63,000 images, each annotated with 6 artifact types.
    \item We present a grounding model tailored to the characteristics of artifacts produced by SR methods, that outperforms other approaches on distortion segmentation tasks.
    \item We release source-disjoint splits, per-mask prominence metadata, inference code, and model weights, and report class-wise crowdsourcing reliability to support reproducible use of the benchmark.
\end{itemize}

The dataset, image quality grounding model inference code and pretrained weights are available at \url{https://huggingface.co/datasets/Divotion/SR-Ground}.

\section{Related Work}

In this section we describe existing IQA datasets and image quality grounding methods.

\subsection{Datasets}

%Single-image SR has benefited from diverse high-resolution datasets designed for both training and evaluation. DIV2K~\cite{Agustsson_2017_CVPR_Workshops} is a widely used benchmark containing 1,000 high-quality 2K images with corresponding low-resolution (LR) images generated via bicubic downsampling at $\times2$, $\times3$, and $\times4$. Flickr2K~\cite{agustsson2017flicker2k}, consisting of 2,650 high-resolution images collected from Flickr, complements DIV2K to form the combined DF2K training set, which has become a standard large-scale corpus for modern SISR models. For evaluation, smaller datasets such as Set5, Set14, BSD100, Urban100, and Manga109 provide challenging scenes covering both synthetic and real-world degradations.

Most IQA datasets provide human-centric perceptual evaluations rather than pixel-wise fidelity metrics. Large-scale in-the-wild datasets such as KonIQ-10k~\cite{koniq10k} and SPAQ~\cite{Fang_2020_CVPR} provide millions of human ratings for diverse authentic distortions. Classic full-reference datasets, including LIVE~\cite{sheikh2006statistical}, CSIQ~\cite{larson2010most}, and PieAPP~\cite{Prashnani_2018_CVPR}, and no-reference datasets such as FLIVE~\cite{ying2019patches} and BIQ2021~\cite{ahmed2022biq2021}, remain standard benchmarks for evaluating perceptual quality metrics.

Recent work has explored grounded IQA, which links perceptual quality scores to explicit spatial annotations. \textbf{QGround}~\cite{chen2024qground} introduces QGround-100K, a large-scale data set that combines 100,000 images with pixel-level distortion masks and textual quality descriptions, supporting explainable IQA. \textbf{GroundingIQA}~\cite{chen2026grounding} further extends this paradigm with GIQA-160K, providing automated bounding-box annotations and detailed textual quality descriptions to enable interpretable, region-aware, and language-guided IQA. While these datasets enable reasoning about distortions beyond scalar quality scores, they are not tailored to SR-specific artifacts, motivating the creation of SR-Ground.

\subsection{Image Quality Grounding Models}
Most existing Image Quality Grounding methods focus on distortion detection rather than segmentation~\cite{Liao_2025_ICCV, chen2026grounding}. In this work, we consider only models capable of segmenting the pixel-level distortion.

\textbf{Q-Ground}~\cite{chen2024qground} addresses the segmentation task, identifying SegFormer and Mask2Former as strong baselines.

\textbf{SegFormer}~\cite{Xie_2021_NeurIPS} is a lightweight segmentation model based on Mix Vision Transformer backbones and a simple decoder that fuses multi-level features with MLPs. Its efficiency and straightforward training make it a popular choice for various segmentation tasks.

\textbf{Mask2Former}~\cite{Cheng_2022_CVPR} provides a universal segmentation framework capable of semantic, instance, and panoptic segmentation. Its masked attention mechanism focuses computation on predicted mask regions, improving both efficiency and boundary quality. These properties make Mask2Former effective for tasks requiring precise distortion segmentation.

\section{Iterative Dataset Curation}

We construct SR-Ground through an iterative pipeline designed to progressively improve both annotation quality and model performance. As illustrated in Figure~\ref{fig:iterative-pipeline}, the pipeline consists of four stages: source image selection, SR data generation, initial annotation, and human-in-the-loop refinement.

We focus on a set of distortion types particularly relevant to super-resolution, including blur, overexposure, low-light conditions, noise, jitter, and \emph{SR-specific artifacts}.

\subsection{Source Image Selection}

To construct a dataset suitable for fine-grained image quality grounding in SR, the source images must satisfy three key requirements: semantic diversity, diversity of real-world distortions, and high visual quality with rich spatial structure. The latter is particularly important, as structurally complex images tend to induce more challenging and diverse artifacts when processed by SR models.

To meet these criteria, we draw images from several large-scale IQA and aesthetics datasets, including AVA~\cite{Murray2012AVA}, Waterloo Exploration~\cite{Ma2017Waterloo}, FLIVE~\cite{ying2019patches}, and KonIQ-10K~\cite{koniq10k}. These datasets collectively provide a wide range of scene types and authentic distortions and do not overlap with Q-Ground-100K~\cite{chen2024qground}, which is later used for model pre-training.

To ensure balanced coverage of both semantic content and distortion characteristics, we represent each image using a combination of complementary features. Specifically, we extract semantic embeddings using CLIP~\cite{Radford2021LearningTV} and perceptual quality embeddings using ARNIQA~\cite{agnolucci2024arniqa}. In addition, following the approach to dataset assessment from \cite{ohtani2024rethinking}, we compute three scalar measures: the Spatial Information (SI)~\cite{svqamethods} as a measure of structural richness, the number of segments produced by SAM~\cite{kirillov2023segany}, which serves as a proxy for spatial complexity, and a blockiness metric~\cite{ohtani2024rethinking}, which captures compression-related artifacts. We then cluster all candidate images into 1,000 groups using K-means with a composite distance metric defined as the average of four normalized distances:

$d(X,Y)=\tfrac{1}{4}\sum_{i\in{\text{ARNIQA},\text{CLIP},\text{SAM},\text{Blockiness}}} d_i(X,Y),$

where cosine distance is used for embedding features and absolute difference for scalar features. All scalar features are normalized to the range $[0,1]$.

From each cluster, we select the image closest to the cluster centroid, resulting in a diverse subset of representative samples. This process yields a final set of 1,000 source images.

\begin{figure}[h!]
\includegraphics[width=\linewidth]{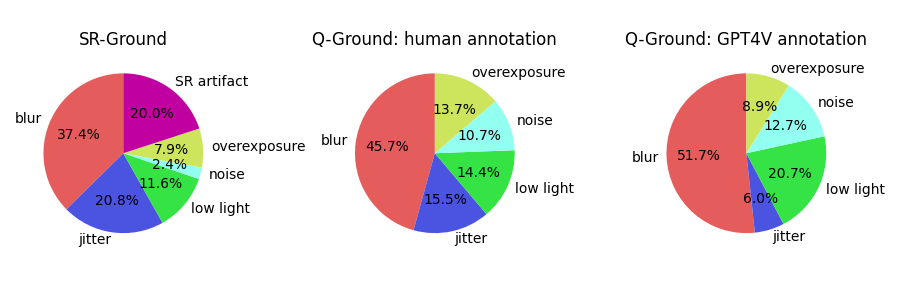}
\caption{Class distribution statistics in the SR-Ground and Q-Ground datasets.}
\Description{Three pie charts comparing the proportions of distortion classes in SR-Ground, manually annotated Q-Ground, and GPT-4V-annotated Q-Ground.}
\label{fig:classes-disribution}
\end{figure}

\subsection{SR Data Generation}

To generate a diverse set of SR inputs, we first construct low-resolution (LR) images using controlled degradation processes. Each source image is downsampled using bicubic interpolation with scaling factors of $2\times$ and $4\times$. To further increase variability in degradation characteristics, we optionally apply Gaussian blurring with kernel sizes of $5 \times 5$ and $9 \times 9$, while also retaining non-blurred versions. This results in 6,000 LR images spanning different levels of detail loss and smoothness.

We then upscale these LR images using a set of state-of-the-art SR methods covering multiple architectural paradigms. We used RealSR~\cite{Ji_2020_CVPR_Workshops}, BSRGAN~\cite{zhang2021designing}, Real-ESRGAN~\cite{wang2021realesrgan}, SwinIR~\cite{liang2021swinir}, ATD~\cite{Zhang_2024_CVPR}, SUPIR~\cite{yu2024scaling} (SUPIR\_Q preset), PASD~\cite{yang2023pasd} (SDXL-based~\cite{ICLR2024_081b0806}), PassionSR~\cite{zhu2025passionsr} (W8A8 version), and DiT4SR~\cite{duan2025dit4sr}.

%The information on the selected models is shown in Table \ref{tab:used-sr-models}.

%\begin{table}[t]
%\centering
%\caption{Super-resolution models used for dataset generation.}
%\label{tab:sr_models}
%\begin{tabular}{l l l}
%\toprule
%\textbf{Category} & \textbf{Model} & \textbf{Configuration} \\
%\midrule
%CNN-based 
%& RealSR~\cite{Ji_2020_CVPR_Workshops} 
%& DF2K, DF2K\_JPEG \\

%\midrule
%GAN-based 
%& BSRGAN~\cite{zhang2021designing} 
%%& Default \\
%& Real-ESRGAN~\cite{wang2021realesrgan} 
%& Default \\

%\midrule
%Transformer-based 
%& SwinIR~\cite{liang2021swinir} 
%& Default; Large \\
%& ATD~\cite{Zhang_2024_CVPR} 
%& Default \\

%\midrule
%Diffusion-based 
%& SUPIR~\cite{yu2024scaling} 
%& SUPIR\_Q \\
%& PASD~\cite{yang2023pasd} 
%& SDXL-based~\cite{ICLR2024_081b0806} \\
%& PassionSR~\cite{zhu2025passionsr} 
%& W8A8 \\
%& DiT4SR~\cite{duan2025dit4sr} 
%& Default \\
%\bottomrule
%\end{tabular}
%\label{tab:used-sr-models}
%\end{table}

These models are selected as representative approaches within their respective subdomains, each known to produce distinct artifact patterns. This diversity is critical for ensuring broad coverage of SR-specific distortions in the final dataset. Applying all SR methods to the generated LR inputs yields a total of 63,000 super-resolved images.

\subsection{Initial Annotation}
To obtain initial pixel-level annotations for SR-generated images, we construct a grounding model capable of segmenting both real-world distortions and SR-specific artifacts. This model serves as the starting point for the iterative refinement pipeline.

\textbf{Base Dataset Preparation}. We consider six distortion categories in SR-Ground: five common real-world distortions (blur, noise, jitter, overexposure, and low-light) adopted from Q-Ground-100K, and an additional category corresponding to SR-specific artifacts, introduced in various papers researching this topic~\cite{desra, molodetskikh2026srprominencecrowdsourcedprotocoldataset}.

The annotator-facing definition follows the operational definition in the Introduction: the SR-artifact mask should cover localized SR-induced failures that cannot be assigned to blur, noise, jitter, overexposure, or low-light. This definition deliberately groups visually different failures, including fabricated texture and structural deformation, into one class rather than imposing an unsupported subtype taxonomy.

To model real-world distortions, we rely on Q-Ground-100K as the primary training source. However, in our setting, which focuses on fine-grained artifact localization for SR outputs, we observe that directly training segmentation models on this dataset leads to occasional training instability. We attribute this behavior to differences between the original Q-Ground task and our setting, as well as to inherent variability in the annotation quality. We observe cases with imprecise mask boundaries, low agreement between annotators, and annotations dominated by a single class covering most of the image.

To improve training stability and label consistency, we introduce a filtering procedure that removes annotations with excessively large dominant regions. Specifically, for each annotation, we compute the normalized area of each class and discard masks where the largest class exceeds a predefined threshold (0.8). This filtering is applied to both manually annotated and GPT-4V-generated subsets of Q-Ground-100K, resulting in a more consistent training set.

To incorporate SR-specific artifacts, we further expand the training data with additional sources, including training split of Open Images dataset~\cite{molodetskikh2026srprominencecrowdsourcedprotocoldataset}. This combined dataset enables joint learning of real-world distortions and SR artifacts.

%\begin{algorithm}
%\caption{Filtering annotations by dominant class proportion}
%\label{alg:qground-filtering-algorithm}
%\begin{algorithmic}[1]

%\Require Dataset $\mathcal{I}$, set of classes $\mathcal{D} = \{d_1, \dots, d_K\}$
%\Ensure Filtered dataset $\mathcal{I}'$

%\For{each object $o_i \in \mathcal{I}$}
%    \State $\text{to\_delete} \gets \emptyset$
    
%    \For{each annotation $a_j \in \text{ann\_list}(o_i)$}
%        \State Load binary mask $M \in \mathbb{R}^{H \times W \times K}$
        
%        \For{$k = 1$ \textbf{to} $K$}
            
%            \State Compute normalized area:
%            \[
%            S_k = \frac{1}{H \cdot W} \sum_{p} M_k(p)
%            \]
%        \EndFor
        
%        \State $S_{\max} \gets \max\limits_{k} S_k$
        
%        \If{$S_{\max} > 0.8$}
%            \State add $j$ to $\text{to\_delete}$
%        \EndIf
%    \EndFor
    
%    \State Remove all annotations with indices in $\text{to\_delete}$
    
%    \If{$|\text{ann\_list}(o_i)| = 0$}
%        \State remove object $o_i$ from $\mathcal{I}$
%    \EndIf
%\EndFor

%\State \Return $\mathcal{I}$

%\end{algorithmic}
%\end{algorithm}

\textbf{Initial Image Quality Grounding Model Training}. We train an Image Quality Grounding model to produce initial artifact segmentation masks. We use two representative segmentation architectures: SegFormer~\cite{Xie_2021_NeurIPS} and Mask2Former~\cite{Cheng_2022_CVPR}. In contrast to prior work, we tested training with both a combination of cross-entropy and Dice losses and a combination of binary cross-entropy and Dice losses, as the former is more standard in the segmentation task.

%We consider two training strategies: (1) a unified approach, where a single model is trained to predict all distortion types from the outset, and (2) a staged approach, where SR artifact segmentation is handled by a separate model (Molodetskikh et al.~\cite{molodetskikh2026srprominencecrowdsourcedprotocoldataset}) and merged later.

Models are trained on different combinations of the prepared datasets and evaluated using mIoU and mAcc on Q-Ground-100K test set. Detailed training configurations and extended results are provided in Appendix~\ref{app:grounding-training}. As shown in Table~\ref{tab:metrics_training_results}, Mask2Former trained on the filtered Q-Ground-100K dataset with the unified multi-class setup provides the strongest balance of the two metrics, with the best mAcc and the second-best mIoU on the filtered Q-Ground-100K test set. Based on these results, we adopt Mask2Former as an initial model for dataset generation.

%During training, a batch size of 40 was used, along with the Adam optimizer with a weight decay of 1e-4. The learning rate for SegFormer was set to 1e-5 for the encoder and 5e-5 for the decoder. The learning rate for Mask2Former was set to 1e-6 for the backbone and 1e-5 for the decoder. A cosine scheduler was also used with a warmup period equal to 10\% of the total number of iterations. Training was performed on 8$\times$NVIDIA A100-82G GPUs for 50,000 iterations.

%Importantly, incorporating SR artifact prediction does not significantly degrade performance on real-world distortions, while achieving competitive results on SR-specific benchmarks.

\begin{table}[]
    \centering
    %\caption{Model training results. In the ``Loss'' column, `BCE' indicates that the model was trained using a combination of BCE loss and Diceloss, while `CE' indicates that the model was trained using a combination of CE loss and Diceloss. In the ``Preset'' column: `U' means that unfiltered Q-Ground-100K was used, `F' means that filtered Q-Ground-100K was used; `M' means that only manually annotated Q-Ground-100K was used, `V' means that Q-Ground-100K annotated by VLM was additionally used.}
    \caption{Model performance on filtered Q-Ground under different training configurations. Higher is better. The best result is \textbf{bolded}, the second best is \underline{underlined}. Extended results are in Appendix~\ref{app:grounding-training}.}
    \setlength{\tabcolsep}{3.5pt}
    \small\begin{tabular}{ccccccc}
        \hline
         %Model & Loss & Filter & Annot. & Resolut. & {mIoU$\uparrow$ / mAcc$\uparrow$} \\
         Model & Filter & Annot. & Resolution & mIoU$\uparrow$ & mAcc$\uparrow$ \\
         \hline
         %SegFormer & BCE & U & M & 448 & .573/.650/.475/.593 \\
         SegFormer & U & M & 448 & .475 & .593 \\  %.573/.650/
         %SegFormer & CE & U & M & 448 & .567/.648/.473/.596 \\
         %SegFormer & BCE & U & V & 448 & .547/.620/.446/.559 \\
         SegFormer & U & V & 448 & .446 & .559 \\ %.547/.620/
         %SegFormer & CE & U & V & 448 & .537/.608/.435/.544 \\
         %SegFormer & BCE & F & M & 448 & .554/.616/.527/.619 \\
         SegFormer & F & M & 448 & .527 & .619 \\ %.554/.616/
         %SegFormer & CE & F & M & 448 & .554/.623/.522/\underline{.623} \\
         %SegFormer & BCE & F & V & 448 & .507/.564/.472/.555 \\
         SegFormer & F & V & 448 & .472 & .555 \\ %.507/.564/
         %SegFormer & CE & F & V & 448 & .509/.564/.465/.546 \\
         %SegFormer & BCE & U & M & 1024 & .530/.625/.411/.557 \\
         SegFormer & U & M & 1024 & .411 & .557 \\ %.530/.625/
         %SegFormer & CE & U & M & 1024 & .526/.620/.405/.550 \\
         %SegFormer & BCE & U & V & 1024 & .504/.578/.380/.495 \\
         SegFormer & U & V & 1024 & .380 & .495 \\ %.504/.578/
         %SegFormer & CE & U & V & 1024 & .485/.555/.358/.466 \\
         %SegFormer & BCE & F & M & 1024 & .528/.594/.486/.585 \\
         SegFormer & F & M & 1024 & .486 & .585 \\ %.528/.594/
         %SegFormer & CE & F & M & 1024 & .524/.588/.473/.570 \\
         %SegFormer & BCE & F & V & 1024 & .443/.495/.400/.475 \\
         SegFormer & F & V & 1024 & .400 & .475 \\ %.443/.495/
         %SegFormer & CE & F & V & 1024 & .459/.510/.405/.481 \\
         \hline
         %Mask2Former & BCE & U & M & 448 & \underline{.586}/.657/.496/.604 \\
         Mask2Former & U & M & 448 & .496 & .604 \\ %\underline{.586}/.657/
         %Mask2Former & CE & U & M & 448 & .553/.625/.454/.563 \\
         %Mask2Former & BCE & U & V & 448 & .539/.606/.435/.537 \\
         Mask2Former & U & V & 448 & .435 & .537 \\ %.539/.606/
         %Mask2Former & CE & U & V & 448 & .512/.586/.402/.514 \\
         %Mask2Former & BCE & F & M & 448 & .559/.619/\underline{.530}/.619 \\
         Mask2Former & F & M & 448 & .530 & .619 \\ %.559/.619/
         %Mask2Former & CE & F & M & 448 & .523/.576/.486/.563 \\
         %Mask2Former & BCE & F & V & 448 & .472/.523/.418/.493 \\
         Mask2Former & F & V & 448 & .418 & .493 \\ %.472/.523/
         %Mask2Former & CE & F & V & 448 & .428/.469/.367/.426 \\
         %Mask2Former & BCE & U & M & 1024 & \textbf{.591}/\textbf{.671}/.498/.621 \\
         Mask2Former & U & M & 1024 & .498 & \underline{.621} \\ %\textbf{.591}/\textbf{.671}/
         %Mask2Former & CE & U & M & 1024 & .580/\underline{.665}/.480/.611 \\
         %Mask2Former & BCE & U & V & 1024 & .539/.609/.426/.534 \\
         Mask2Former & U & V & 1024 & .426 & .534 \\ %.539/.609/
         %Mask2Former & CE & U & V & 1024 & .511/.573/.392/.486 \\
         %Mask2Former & BCE & F & M & 1024 & .564/.630/\textbf{.534}/\textbf{.632} \\
         Mask2Former & F & M & 1024 & \underline{.534} & \textbf{.632} \\ %.564/.630/
         %Mask2Former & CE & F & M & 1024 & .535/.593/.496/.583 \\
         %Mask2Former & BCE & F & V & 1024 & .496/.549/.463/.540 \\
         Mask2Former & F & V & 1024 & .463 & .540 \\ %.496/.549/
         %Mask2Former & CE & F & V & 1024 & .479/.529/.437/.511 \\
         \hline
         EoMT-DINOv3 & F & M & 512 & \textbf{.538} & .620 \\
         OneFormer & F & M & 512 & .529 & .618 \\
         \hline
    \end{tabular}
    \label{tab:metrics_training_results}

    \vspace{2mm}
    \begin{flushleft}
    \footnotesize
    %Loss: BCE = BCE + Dice, CE = CE + Dice. \\
    Filter: U = unfiltered, F = filtered. \\
    Annot.: M = manual, V = VLM (GPT4V).
    \end{flushleft}
\end{table}

We screened EoMT-DINOv3~\cite{kerssies2025eomt} and OneFormer~\cite{jain2023oneformer} at this initial Q-Ground stage using the filtered, manually annotated subset at \(512\times512\) resolution. Table~\ref{tab:metrics_training_results} reports only their best checkpoints; versions trained with the common pipeline used for Mask2Former and SegFormer performed worse. Neither additional architecture consistently improves on Mask2Former: EoMT-DINOv3 raises mIoU only from .534 to .538 while reducing mAcc from .632 to .620, and OneFormer reaches .529 mIoU and .618 mAcc. We therefore screened out both models before the SR-Ground and Open Images experiments. The remaining experiments use Mask2Former as the higher-accuracy high-resolution model and SegFormer as the faster alternative, preserving the two standard baselines used in Q-Ground~\cite{chen2024qground}.

\begin{figure*}[h!]
\includegraphics[width=0.9\linewidth]{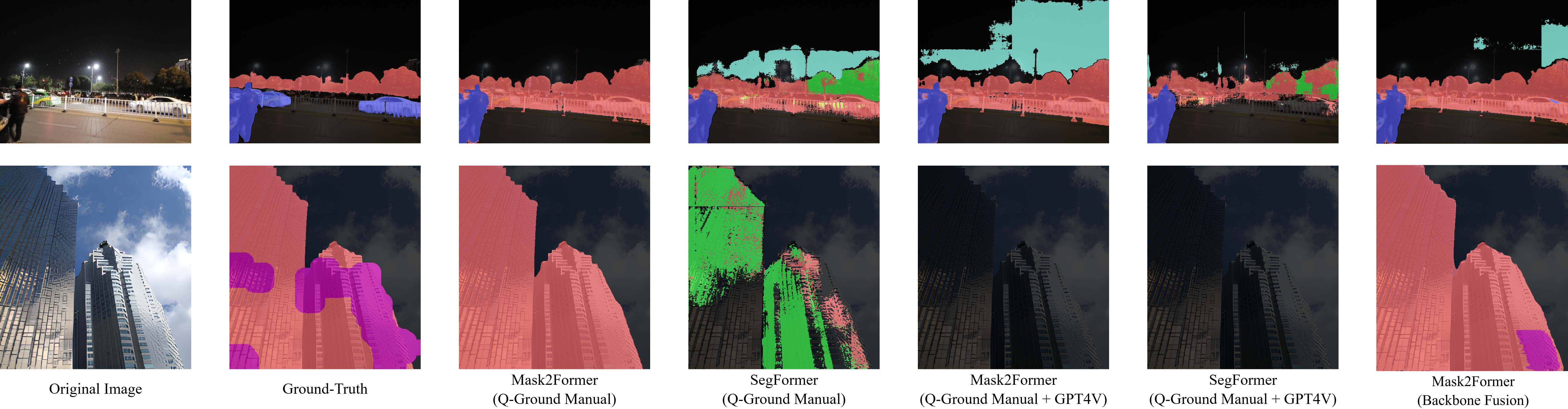}
\caption{Comparison of Image Quality Grounding methods on samples from Q-Ground (top) and SR-Ground (bottom). Distortions: \hlc[jitter-color]{jitter}, \hlc[noise-color]{noise}, \hlc[overexposure-color]{overexposure}, \hlc[blur-color]{blur}, \hlc[lowlight-color]{low light}, \hlc[srartifact-color]{SR artifact}.}
\Description{Grid of natural distorted and super-resolved images with color-coded masks predicted by the compared grounding models.}
\label{fig:model-visual-comparison}
\end{figure*}

\subsection{Human-in-the-Loop Refinement}

After applying our initial model to super-resolved images, we obtain the first iteration of SR-Ground annotations. Since the model is trained on Q-Ground, it is limited to detecting only the distortion classes defined in that dataset. To account for \emph{SR-specific artifacts}, we additionally leverage the method proposed in \cite{molodetskikh2026srprominencecrowdsourcedprotocoldataset}.

To further improve annotation quality and incorporate human judgment, we introduce a human-in-the-loop refinement stage. This is implemented via a crowdsourced annotation process on the \href{https://tasks.yandex.com/}{Yandex Tasks} platform. Each annotator is presented with a pair of images: the SR-processed image and its corresponding LR version upscaled via bicubic interpolation. For each highlighted mask, the annotator indicates whether the SR image contains a visible distortion of the selected type within the highlighted area. The interface of crowdsourcing service and extended details of subjective experiments are described on the dataset page. %is shown in Figure~\ref{fig:yandex-tasks-interface}.

%\begin{figure}[h!]
%\includegraphics[width=1.0\linewidth]{samples/img/yandex-tasks-interface.png}
%\caption{Yandex Tasks interface. SR artifact assessment task.}
%\label{fig:yandex-tasks-interface}
%\end{figure}

We quantify the \textit{prominence} of each artifact as the fraction of annotators who confirm its presence. A pilot study with 40 votes per mask showed that reducing the final budget from 30 to 10 votes increases the average 95\% confidence-interval length by only 0.075 while reducing annotation cost by a factor of three. We therefore collect 10 votes per selected mask. The resulting prominence rescales the corresponding per-pixel class probabilities, after which each pixel is assigned the class with the largest adjusted probability. Crowdsourced refinement is retained for blur, low-light, overexposure, and SR artifacts. Pilot judgments for jitter and noise were unreliable, so these two classes retain their model-generated annotations.

\begin{table}[t]
\centering
\caption{Crowdsourcing agreement by refined class. ``Outside'' uses prominence in $\leq 0.4$ or $\geq 0.6$; ``strong'' uses prominence $\leq 0.2$ or $\geq 0.8$.}
\label{tab:annotation-agreement}
\setlength{\tabcolsep}{3.2pt}
\tiny
\begin{tabular}{lcccc}
\hline
Class & Pairwise & Outside & Strong & Kripp.\ $\alpha$ \\
\hline
Blur & .57 & .64 & .77 & .14 \\
Low-light & .86 & .89 & .92 & .24 \\
Overexposure & .62 & .69 & .76 & .23 \\
SR artifact & .68 & .73 & .83 & .13 \\
\hline
\end{tabular}
\end{table}

As shown in Table~\ref{tab:annotation-agreement}, agreement increases for all four classes after excluding the ambiguous prominence band and increases further for strong masks. The modest chance-corrected agreement is expected because masks were selected specifically where the model was uncertain; retaining prominence rather than forcing a binary consensus records this uncertainty.

This refined dataset is then used to fine-tune the grounding model, producing a stronger model for both real-world distortions and SR artifacts. The procedure can be repeated iteratively:

\begin{table*}[t]
\centering
\caption{Performance on the filtered Q-Ground-100K test set. All datasets specified in the ``Training Data'' column have been filtered. The best result is \textbf{bolded}, the second best is \underline{underlined} within each architecture or training-data block.}
\label{tab:qground_results}
\tiny\begin{tabular}{l|l|l|c|c|c|c|c|c|c|c|c|c|c|c}
\hline
\multirow{2}{*}{Model} & \multirow{2}{*}{Training Data} & \multirow{2}{*}{Training Strategy} & \multicolumn{2}{|c|}{\cellcolor{jitter-color!75}jitter} & \multicolumn{2}{|c|}{\cellcolor{noise-color!75}noise} & \multicolumn{2}{|c|}{\cellcolor{overexposure-color!75}overexposure} & \multicolumn{2}{|c|}{\cellcolor{blur-color!75}blur} & \multicolumn{2}{|c|}{\cellcolor{lowlight-color!75}low light} & \multicolumn{2}{|c}{Average} \\\cline{4-15}
& & & mIoU $\uparrow$ & mAcc $\uparrow$ & mIoU $\uparrow$ & mAcc $\uparrow$ & mIoU $\uparrow$ & mAcc $\uparrow$ & mIoU $\uparrow$ & mAcc $\uparrow$ & mIoU $\uparrow$ & mAcc $\uparrow$ & mIoU $\uparrow$ & mAcc $\uparrow$ \\
\hline
Baseline (SegFormer) & Q-Ground (real) & From Scratch & \underline{0.622} & \underline{0.717} & \textbf{0.223} & \textbf{0.289} & 0.469 & 0.537 & 0.477 & 0.592 & \textbf{0.482} & \textbf{0.600} & 0.486 & 0.585 \\
SegFormer & Q-Ground (GPT4V) & Fine-tuned Baseline & 0.001 & 0.001 & 0.028 & 0.029 & 0.350 & 0.398 & 0.181 & 0.237 & 0.212 & 0.263 & 0.158 & 0.193 \\
SegFormer & Q-Ground (real + GPT4V) & Fine-tuned Baseline & 0.443 & 0.497 & 0.195 & 0.252 & 0.466 & \underline{0.561} & 0.407 & 0.492 & 0.367 & 0.421 & 0.399 & 0.470 \\
SegFormer & Q-Ground (real) + SR-Ground & From Scratch & \textbf{0.630} & 0.708 & \underline{0.217} & 0.273 & 0.474 & 0.547 & 0.496 & 0.602 & \underline{0.446} & \underline{0.542} & \underline{0.489} & 0.576 \\
SegFormer & Q-Ground (real) + SR-Ground & Fine-tuned Baseline & \underline{0.622} & \underline{0.717} & 0.212 & \underline{0.278} & \underline{0.491} & \textbf{0.583} & \underline{0.503} & \underline{0.621} & 0.428 & 0.509 & \underline{0.489} & \underline{0.586} \\
SegFormer & SR-Ground & Fine-tuned Baseline & \textbf{0.630} & \textbf{0.731} & 0.188 & 0.231 & \textbf{0.495} & \textbf{0.583} & \textbf{0.577} & \textbf{0.800} & 0.355 & 0.414 & \textbf{0.503} & \textbf{0.631} \\
\hline
Baseline (Mask2Former) & Q-Ground (real) & From Scratch & \textbf{0.727} & \textbf{0.800} & 0.152 & 0.179 & 0.498 & 0.575 & \underline{0.573} & \underline{0.727} & 0.436 & \underline{0.517} & \textbf{0.534} & \underline{0.632} \\
Mask2Former & Q-Ground (GPT4V) & Fine-tuned Baseline & 0.000 & 0.000 & 0.019 & 0.020 & 0.343 & 0.397 & 0.250 & 0.342 & 0.177 & 0.201 & 0.174 & 0.218 \\
Mask2Former & Q-Ground (real + GPT4V) & Fine-tuned Baseline & 0.445 & 0.471 & 0.144 & 0.181 & 0.480 & 0.559 & 0.384 & 0.453 & 0.406 & 0.465 & 0.395 & 0.451 \\
Mask2Former & Q-Ground (real) + SR-Ground & From Scratch & 0.706 & 0.774 & \textbf{0.209} & \textbf{0.278} & \textbf{0.520} & \textbf{0.616} & 0.505 & 0.622 & \textbf{0.450} & \textbf{0.540} & 0.517 & 0.610 \\
Mask2Former & Q-Ground (real) + SR-Ground & Fine-tuned Baseline & \underline{0.714} & \underline{0.780} & \underline{0.173} & \underline{0.215} & \underline{0.518} & \underline{0.609} & 0.537 & 0.661 & \underline{0.439} & 0.512 & \underline{0.524} & 0.613 \\
Mask2Former & SR-Ground & Fine-tuned Baseline & 0.691 & 0.755 & 0.146 & 0.177 & 0.508 & 0.602 & \textbf{0.580} & \textbf{0.814} & 0.351 & 0.406 & 0.515 & \textbf{0.638} \\
%\hline

%Mask2Former & \makecell[l]{Q-Ground (real) \\ + Open Images~\cite{molodetskikh2026srprominencecrowdsourcedprotocoldataset} (SR artifacts)} & From Scratch & 0.630 & 0.713 & \textbf{0.159} & \textbf{0.197} & 0.399 & 0.453 & 0.487 & \textbf{0.637} & 0.324 & 0.411 & 0.448 & 0.545 \\
%Mask2Former & \makecell[l]{Q-Ground (real) \\ + Open Images~\cite{molodetskikh2026srprominencecrowdsourcedprotocoldataset} (SR artifacts) \\ + SR-Ground (SR artifacts included)}& From Scratch & \underline{0.687} & \underline{0.759} & \underline{0.153} & \underline{0.179} & \textbf{0.507} & \textbf{0.601} & \underline{0.506} & \textbf{0.637} & \underline{0.450} & \textbf{0.547} & \underline{0.506} & \textbf{0.601} \\
%Mask2Former & \makecell[l]{Q-Ground (real) \\ + Open Images~\cite{molodetskikh2026srprominencecrowdsourcedprotocoldataset} (SR artifacts) \\ + SR-Ground (SR artifacts included)} & Fine-tuned Baseline & \textbf{0.703} & \textbf{0.777} & 0.136 & 0.157 & \underline{0.504} & \underline{0.585} & \textbf{0.512} & \underline{0.629} & \textbf{0.452} & \underline{0.536} & \textbf{0.509} & \underline{0.596} \\
\hline
\end{tabular}
\end{table*}

\begin{figure*}[h!]
\includegraphics[width=0.85\linewidth]{samples/img/architectures.pdf}
\caption{Mask2Former architectures considered in our experiments: (a) the default no-reference model; (b) Input Fusion; (c) Decoder Fusion; and (d) Backbone Fusion.}
\Description{Four Mask2Former variants: a no-reference baseline and three reduced-reference fusion architectures operating at the input, pixel-decoder, or backbone levels.}
\label{fig:mask2former_archs}
\end{figure*}

\begin{table*}[t]
\centering
\caption{Results of some open-source SR artifact segmentation models and proposed image quality grounding models on SR artifact detection datasets. Extended results are presented on the DeSRA dataset. The best result is \textbf{bolded}, the second best is \underline{underlined}. Avg. Rank is the mean rank across the four SRCC datasets; unavailable entries are omitted.}
\label{tab:sr_results}
\tiny\begin{tabular}{l|c|c|c|c|c|c|c|c|c|c|c|c}
    \hline
    \multirow{2}{*}{} &
    \multirow{2}{*}{Needs Reference} &
    \multirow{2}{*}{\makecell{SRCC \\ Open Images}$\uparrow$} &
    \multirow{2}{*}{\makecell{SRCC \\ Urban100}$\uparrow$} &
    \multirow{2}{*}{\makecell{SRCC \\ Urban100-HR}$\uparrow$} &
    \multirow{2}{*}{\makecell{SRCC \\ DeSRA}$\uparrow$} &
    \multirow{2}{*}{\makecell{Avg. Rank \\ SRCC}$\downarrow$} &
    \multicolumn{4}{c|}{DeSRA} &
    \multicolumn{2}{c}{Q-Ground} \\
    \cline{8-13}
    & & & & & & & \makecell{F1 \\ (Full)}$\uparrow$ & \makecell{PR-AUC \\ (Full)}$\uparrow$ & \makecell{IoU \\ (Prominent)}$\uparrow$ & \makecell{PR-AUC \\ (Prominent)}$\uparrow$ & mIoU$\uparrow$ & mAcc$\uparrow$ \\
    \hline
    TOPIQ~\cite{chen2024topiq} & $\times$ & -0.096 & -0.099 & -0.015 & -0.114 & 15.75 & 0.016 & 0.003 & 0.042 & 0.107 & - & - \\
    PaQ-2-PiQ~\cite{ying2019patches} & $\times$ & -0.097 & -0.222 & 0.024 & -0.088 & 15.50 & 0.016 & 0.001 & 0.031 & 0.029 & - & - \\
    ERQA~\cite{kirillova2022erqa} & \ding{51} & -0.134 & 0.138 & 0.025 & 0.023 & 13.50 & 0.040 & 0.003 & 0.051 & 0.009 & - & - \\
    PAL4VST~\cite{Zhang_2023_ICCV} & $\times$ & 0.136 & 0.145 & N/A & 0.043 & 11.33 & 0.005 & N/A & 0.014 & N/A & - & - \\
    PAL4Inpaint~\cite{zhang2022perceptual} & $\times$ & 0.067 & 0.156 & 0.112 & 0.109 & 11.25 & 0.061 & N/A & 0.114 & N/A & - & - \\
    DeSRA~\cite{desra} & \ding{51} & 0.299 & 0.049 & 0.340 & 0.021 & 10.00 & 0.170 & 0.055 & \underline{0.508} & \textbf{0.661} & - & - \\
    bd\_jup~\cite{molodetskikh2026srprominencecrowdsourcedprotocoldataset} & \ding{51} & 0.177 & 0.245 & 0.267 & -0.098 & 9.75 & 0.118 & 0.024 & 0.363 & 0.177 & - & - \\
    LPIPS~\cite{zhang2018perceptual} & \ding{51} & 0.304 & 0.234 & \textbf{0.446} & -0.132 & 8.25 & 0.146 & 0.039 & 0.409 & 0.501 & - & - \\
    ssm\_jup~\cite{molodetskikh2026srprominencecrowdsourcedprotocoldataset} & \ding{51} & 0.237 & 0.099 & 0.099 & \textbf{0.342} & 9.50 & 0.169 & 0.035 & 0.393 & 0.365 & - & - \\
    LDL~\cite{jie2022LDL} & \ding{51} & 0.233 & 0.200 & 0.255 & 0.193 & 8.75 & 0.162 & 0.050 & 0.344 & 0.418 & - & - \\
    SSIM~\cite{1284395} & \ding{51} & 0.297 & 0.264 & 0.348 & -0.010 & 7.75 & 0.179 & 0.055 & \textbf{0.524} & 0.605 & - & - \\
    DISTS~\cite{ding2020iqa} & \ding{51} & 0.370 & 0.283 & \underline{0.401} & -0.009 & 5.13 & 0.164 & 0.046 & 0.502 & 0.543 & - & - \\
    Molodetskikh et al.~\cite{molodetskikh2026srprominencecrowdsourcedprotocoldataset} & \ding{51} & \underline{0.374} & 0.241 & 0.394 & 0.213 & \underline{4.25} & \underline{0.190} & \textbf{0.061} & 0.505 & 0.603 & - & - \\
    \hline
    \makecell{Ours \\ (No-Reference)} & $\times$ & 0.353 & 0.178 & 0.185 & 0.211 & 7.50 & 0.161 & 0.034 & 0.388 & 0.335 & \textbf{0.526} & \underline{0.620} \\
    \makecell{Ours \\ (Input Fusion)} & \ding{51} & 0.370 & 0.240 & 0.165 & 0.253 & 6.13 & 0.169 & 0.038 & 0.392 & 0.335 & 0.524 & \underline{0.620} \\
    \makecell{Ours \\ (Decoder Fusion)} & \ding{51} & 0.339 & \underline{0.316} & 0.184 & 0.245 & 5.50 & 0.186 & \underline{0.058} & 0.369 & 0.400 & \underline{0.525} & \textbf{0.624} \\
    \makecell{Ours \\ (Backbone Fusion)} & \ding{51} & \textbf{0.395} & \textbf{0.406} & 0.397 & \underline{0.311} & \textbf{1.75} & \textbf{0.193} & 0.050 & 0.489 & \underline{0.627} & 0.512 & 0.597 \\
    \hline
\end{tabular}
\end{table*}

\begin{enumerate}
    \item Select new SR images from the original source dataset.
    \item Annotate them using the current model.
    \item Refine masks via crowdsourcing.
    \item Fine-tune the current model on the updated dataset
\end{enumerate}

Through this iterative process, the model and the dataset co-evolve, progressively improving both segmentation accuracy and the reliability of artifact localization. We annotated SR-Ground through three iterations of the proposed refinement, resulting in a dataset of 63,000 images containing grounding masks for 6 distortion types. Figure~\ref{fig:classes-disribution} shows the distribution of classes in Q-Ground and SR-Ground.

\section{Experiments}
\subsection{Training Grounding Models}

We evaluate SR-Ground by fine-tuning state-of-the-art grounding models and comparing them with models trained on Q-Ground-100K. We consider two architectures: SegFormer and Mask2Former. Each model uses six output channels corresponding to the target distortion types, including \textit{SR-specific artifacts}. When training on Q-Ground-100K, only five classes are supervised, and the SR artifact channel is ignored (set to zero). Full six-class supervision is applied only when training on SR-Ground.

All models are initialized by training on the manually annotated filtered training subset of Q-Ground-100K. To study the effect of synthetic data, we apply two fine-tuning strategies. In the first, models are further trained on the synthetic subset of Q-Ground-100K. In the second, the same pretrained models are fine-tuned (or trained from scratch) on SR-Ground.

We evaluate all models on the Q-Ground-100K test set using standard segmentation metrics (mIoU and mAcc) to assess transfer to real-world distortions. The results are reported in Table~\ref{tab:qground_results}. Fine-tuning on SR-Ground slightly improves SegFormer on both average metrics, whereas the corresponding Mask2Former changes are mixed. Thus, the experiment demonstrates transfer to this particular real-distortion test set, but neither uniform improvement across architectures nor universal generalization. %Also, fine-tuning on SR-Ground brought SegFormer to almost the same level of performance as Mask2Former.

Finally, we evaluate the ability of models trained on SR-Ground to detect SR-specific artifacts. For this task, we used both No-Reference (NR) and Reduced-Reference (RR) approaches. We developed three methods for transforming Mask2Former into a Reduced-Reference model. 
Input Fusion combines the SR image, the upsampled LR image, and their residual before the original Mask2Former. Decoder Fusion combines the shared-weight SR and LR multiscale features immediately before the transformer decoder. Backbone Fusion combines shared-weight SR and LR features after each Swin-L backbone~\cite{Liu_2021_ICCV} stage. These approaches are shown in Figure~\ref{fig:mask2former_archs}. We used the low-resolution image upscaled with RLFN~\cite{9857179} as reference for Reduced-Reference approaches.

We use two additional losses for joint real-distortion and SR-artifact segmentation:
\begin{enumerate}
    \item \textbf{KD loss.} BCE distillation from a frozen Mask2Former trained on the five real-distortion classes. It reduces catastrophic forgetting and has weight 0.1.
    \item \textbf{Identity loss.} BCE between zero targets and the SR-artifact logits obtained when the same reference image is supplied as both the SR input and the reference. It encourages the model to distinguish artifacts from LR-consistent content and has weight 0.05.
\end{enumerate}

To evaluate models on SR-artifact segmentation task, we followed the testing methodology introduced by Molodetskikh et al.~\cite{molodetskikh2026srprominencecrowdsourcedprotocoldataset}. We report threshold-free prominence score (marked as ``SRCC'' in all results tables) on four datasets: Open Images, Urban100, Urban100-HR, and DeSRA. We also calculated F1 and PR-AUC on full DeSRA dataset as well as IoU and PR-AUC on its prominent subset, following another Molodetskikh et al. paper~\cite{molodetskikh2026prominenceawareartifactdetectiondataset}. Additional Open Images metrics were not calculated because the compared studies use different versions of that dataset. All the results are shown in Table~\ref{tab:sr_results}. Training on SR-Ground provides a good trade-off between detecting real distortions and SR-generated artifacts. Among the Reduced-Reference approaches, Backbone Fusion performs best in 
SR-artifact segmentation task, achieving the highest values for threshold-free prominence score on Open Images and Urban100, the second highest value on DeSRA and the third highest value on Urban100-HR. It also achieves the best F1 score on the full DeSRA dataset and the second-best PR-AUC on the prominent subset. However, Backbone Fusion exhibits the most significant forgetting of the task of grounding real distortions among all the proposed models.

Figure~\ref{fig:model-visual-comparison} presents visual comparisons of model predictions. Models trained on SR-Ground produce more precise and semantically consistent masks, particularly for SR-specific artifacts. The Reduced-Reference model localizes SR artifacts while retaining useful real-distortion segmentation maps.

\begin{table*}[t]
\centering
\caption{Results on the unseen-SR evaluation benchmark using the matched RLFN-reference protocol and the external-plus-SAM candidate pool. We report 95\% confidence intervals for overall SRCC and AP. The best result is \textbf{bolded}, the second best is \underline{underlined}.}
\label{tab:srground_validation}
\tiny\begin{tabular}{l|c|c|c|c|c|c|c|c|c|c|c|c}
    \hline
    \multirow{2}{*}{Method} &
    \multicolumn{2}{c|}{DIV2K} &
    \multicolumn{2}{c|}{RealSR} &
    \multicolumn{2}{c|}{RealLQ250} &
    \multicolumn{2}{c|}{RealLR200} &
    \multicolumn{2}{c|}{Overall} &
    \multirow{2}{*}{\makecell{Avg. Rank \\ SRCC}$\downarrow$} & 
    \multirow{2}{*}{\makecell{Avg. Rank \\ AP}$\downarrow$}
    \\
    \cline{2-11}
    & SRCC$\uparrow$ & AP$\uparrow$ & SRCC$\uparrow$ & AP$\uparrow$ & SRCC$\uparrow$ & AP$\uparrow$ & SRCC$\uparrow$ & AP$\uparrow$ & SRCC$\uparrow$ & AP$\uparrow$ & &  \\
    \hline
    DeSRA~\cite{desra} & 0.167 & 0.600 & 0.186 & 0.545 & 0.050 & 0.478 & 0.114 & 0.480 & 0.144, [0.092, 0.196] & 0.532, [0.479, 0.586] & 7.75 & 7.00 \\
    DISTS~\cite{ding2020iqa} & 0.253 & 0.620 & 0.445 & \textbf{0.691} & 0.230 & 0.527 & 0.419 & 0.589 & 0.351, [0.296, 0.403] & 0.603, [0.547, 0.663] & 4.25 & 4.00 \\
    SSIM~\cite{1284395} & \underline{0.291} & \underline{0.631} &  \underline{0.451} & 0.680 & 0.235 & 0.517 & 0.417 & 0.595 & \underline{0.367}, [0.317, 0.416] & 0.612, [0.554, 0.668] & 3.75 & 4.25 \\
    Molodetskikh et al.~\cite{molodetskikh2026srprominencecrowdsourcedprotocoldataset} & \textbf{0.342} & \textbf{0.649} & \textbf{0.476} & \underline{0.689} & \underline{0.296} & 0.568 & \underline{0.468} & \underline{0.653} & \textbf{0.407}, [0.356, 0.454] & \textbf{0.640}, [0.580, 0.697] & \textbf{1.50} & \textbf{2.00} \\
    \hline
    Ours (No-Reference) & 0.198 & 0.591 & 0.338 & 0.626 & 0.280 & \textbf{0.613} & 0.308 & 0.606 & 0.284, [0.229, 0.338] & 0.606, [0.549, 0.664] & 5.50 & 4.00 \\
    Ours (Backbone Fusion w/o SR-Ground) & 0.205 & 0.576 & 0.375 & 0.633 & 0.255 & 0.547 &  0.431 & 0.622 & 0.313, [0.250, 0.371] & 0.593, [0.534, 0.650] & 4.00 & 4.75 \\
    Ours (Backbone Fusion, shuffled low-res) & 0.167 & 0.557 & 0.267 & 0.547 & 0.237 & 0.561 & 0.347 & 0.569 & 0.253, [0.199, 0.306] & 0.555, [0.501, 0.610] & 6.50 & 6.50 \\
    Ours (Backbone Fusion) &  0.236 & 0.600 & 0.362 & 0.625 & \textbf{0.336} & \underline{0.601} & \textbf{0.474} & \textbf{0.686} & 0.337, [0.275, 0.395] & \underline{0.617}, [0.559, 0.672] & \underline{2.75} & \underline{3.50} \\
    \hline
\end{tabular}
\end{table*}

\begin{table*}[t]
\centering
\caption{Paired comparisons on the unseen-SR evaluation benchmark. We report 95\% confidence intervals for $\Delta$SRCC and $\Delta$AP.}
\label{tab:iqg_comparison}
\tiny\begin{tabular}{c|c|c|c|c|c|c}
    \hline
    Comparison &
    SRCC A &
    SRCC B &
    $\Delta$ SRCC & 
    AP A &
    AP B &
    $\Delta$ AP
    \\
    \hline
    \makecell{A. Backbone Fusion \\ B. No-Reference} & 0.337 & 0.284 & +0.053, [0.010, 0.095] & 0.617 & 0.606 & +0.011, [-0.015, 0.038] \\
    \hline
    \makecell{A. Backbone Fusion \\ B. Backbone Fusion w/o SR-Ground} & 0.337 & 0.313 & +0.024, [-0.008, 0.056] & 0.617 & 0.593 & +0.023, [-0.003, 0.048] \\
    \hline
    \makecell{A. Backbone Fusion \\ B. Backbone Fusion, shuffled low-res} & 0.337 & 0.253 & +0.084, [0.031, 0.139] & 0.617 & 0.555 & +0.061, [0.028, 0.095] \\
    \hline
    \makecell{A. Backbone Fusion \\ B. Molodetskikh et al.} & 0.337 & 0.407 & -0.070, [-0.114, -0.027] & 0.617 & 0.640 & -0.024, [-0.056, 0.008] \\
    \hline
    \makecell{A. No-Reference \\ B. Backbone Fusion, shuffled low-res} & 0.284 & 0.253 & +0.031, [-0.026, 0.088] & 0.606 & 0.555 & +0.050, [0.017, 0.084] \\
    \hline
\end{tabular}
\end{table*}

\subsection{Unseen-SR Evaluation Benchmark}
Q-Ground provides a test set for real-distortion segmentation, but it does not evaluate transfer to unseen SR methods. We therefore constructed a separate unseen-SR benchmark that is source-disjoint from the SR-Ground training split.

We selected four source datasets:
\begin{enumerate}
    \item \textbf{DIV2K Validation Set}~\cite{Agustsson_2017_CVPR_Workshops}. It is commonly used to evaluate SR models and contains diverse HR images with bicubically downsampled LR counterparts.
    \item \textbf{RealSR v3}~\cite{cai2019ntire}. It is a widely used real-world SR dataset containing spatially complex photographs captured with different cameras.
    \item \textbf{RealLR200}~\cite{wu2024seesr} and \textbf{RealLQ250}~\cite{ai2024dreamclear}. They contain diverse, challenging LR images gathered from previous studies and practical scenarios. Their key distinction is that they do not provide HR ground truth. RealLR200 covers multiple resolutions and aspect ratios, whereas RealLQ250 contains more complex distortions at a fixed resolution.
\end{enumerate}

We selected 50 source images from each dataset. For DIV2K, we did not use the provided bicubic LR images; instead, we generated LR inputs from the HR images with three Real-ESRGAN degradation presets~\cite{wang2021realesrgan}: mild for 17 images, medium for 17, and severe for 16. We then processed all 200 LR inputs with five SR methods absent from both the SR-Ground training split and the datasets evaluated by Molodetskikh et al.~\cite{molodetskikh2026srprominencecrowdsourcedprotocoldataset}:
\begin{enumerate}
    \item \textbf{HAT}~\cite{chen2023activating}. A transformer-based SR method.
    \item \textbf{ResShift}~\cite{yue2023resshift}. A four-step diffusion-based SR method without a text-to-image backbone.
    \item \textbf{SeeSR}~\cite{wu2024seesr}. A 50-step semantic-aware diffusion SR method based on Stable Diffusion 2~\cite{Rombach_2022_CVPR}.
    \item \textbf{VOSR}~\cite{wu2026vosr}. A modern 25-step vision-only diffusion SR method.
    \item \textbf{ODTSR}~\cite{fang2025onestep}. A modern one-step diffusion SR method based on Qwen-Image.
\end{enumerate}

We generated SR-artifact maps using LDL~\cite{jie2022LDL}, DeSRA~\cite{desra}, DISTS~\cite{ding2020iqa}, LPIPS~\cite{zhang2018perceptual}, SSIM~\cite{1284395}, ssm\_jup and bd\_jup~\cite{molodetskikh2026srprominencecrowdsourcedprotocoldataset}, and Molodetskikh et al.~\cite{molodetskikh2026srprominencecrowdsourcedprotocoldataset}. For DIV2K and RealSR, where HR ground truth is available, we ran reference-based methods with both the HR image and the LR image upscaled by RLFN~\cite{9857179}. We also used SAM~\cite{kirillov2023segany} to obtain up to five additional candidate masks per SR output. We evaluated the following proposed models:
\begin{enumerate}
    \item No-Reference model.
    \item Reduced-Reference model with Backbone Fusion.
    \item Backbone-Fusion RR model with shuffled LR references.
    \item Backbone-Fusion RR model trained without SR-Ground, using Q-Ground and Open Images only.
\end{enumerate}
For our models, we applied a sigmoid to the SR-artifact logits to obtain probability maps.

As in the original SR-Ground pipeline, we used crowdsourcing to estimate SR-artifact prominence for every candidate mask covering at least 1\% of the image. The resulting evaluation contains 1,000 SR outputs from 200 source images.

We follow the threshold-free prominence protocol of Molodetskikh et al.~\cite{molodetskikh2026srprominencecrowdsourcedprotocoldataset}. For each candidate region, the method score is difference between the median heatmap value inside the mask and the median value outside it; SRCC measures its rank correlation with human prominence. We also report AP after defining masks with prominence at least 0.5 as positive. The main table uses a common pool of masks proposed by external methods and SAM, excluding masks proposed by the evaluated NR/RR models. We compute 95\% confidence intervals with 5,000 source-level bootstrap resamples, keeping all five SR outputs and their masks together. Table~\ref{tab:srground_validation} shows that Backbone Fusion attains the highest observed SRCC on RealLQ250 and the highest SRCC and AP on RealLR200, the two native real-world datasets without pristine HR references. It also ranks second by overall AP.

Table~\ref{tab:iqg_comparison} separates the effects of the reference and the training data. Backbone Fusion significantly improves SRCC over the NR model and significantly degrades when the LR references are shuffled, showing that it exploits paired LR--SR information rather than additional capacity alone. Adding SR-Ground improves both SRCC and AP, although the paired confidence intervals on the full external proposal pool include zero. Molodetskikh et al. remains strongest on the aggregate benchmark, whereas Backbone Fusion achieves the highest observed performance on the no-HR real-world subset. These results indicate that paired datasets favor generic reference-difference methods, while SR-Ground-trained RR segmentation is particularly effective on native real-world LR inputs.

\section{Grounding-guided Fine-Tuning for Interactive Super-Resolution}
\label{sec:osediff-application}

We use the trained image quality grounding model as a frozen component in a mask-conditioned fine-tuning pipeline for OSEDiff~\cite{wu2024one}. Full training pipeline is described in Appendix~\ref{app:osediff}. The Stable Diffusion 2.1 VAE encoder~\cite{Rombach_2022_CVPR} is augmented with six mask channels. A value of \(-1\), \(+1\), or \(0\) in each channel requests distortion removal, addition, or no edit in the corresponding region. The new columns of the VAE input convolution are zero-initialized and trained directly, while its RGB columns and LoRA factors are frozen; the remaining pretrained OSEDiff LoRA parameters continue training with a reduced learning rate. Thus, a zero control map preserves the original OSEDiff behavior, while the control signal is not constrained by the rank-4 bottleneck of the pretrained LoRA.

Training uses two passes. The first produces an unconditioned reconstruction \(H_0=G_\theta(x_{\mathrm{LQ}},0)\), where $x_{\mathrm{LQ}}$ is a low-quality image obtained with Real-ESRGAN degradation pipeline~\cite{wang2021realesrgan}. The frozen grounding model predicts distortion maps, and SAM~\cite{kirillov2023segany} supplies candidate regions from which two or three distinct mask instructions are sampled. The binary SAM masks are retained for assigning and verifying each command, while linearly smoothed signed maps $C$ are passed to the generator. A second, mask-conditioned pass produces a raw output \(R_1=G_\theta(H_0,C)\). With \(S=\max_k|C_k|\), where $k$ is one of the distortions, the final interactive result is
\[
H_1=H_0+S\odot(R_1-H_0),
\]
which makes the output exactly equal to \(H_0\) outside the soft command support.

The first pass retains the original OSEDiff L1, LPIPS, VSD, and regulator diffusion objectives. The second pass is not supervised against the dataset Ground-Truth and does not use mask-blind VSD. Instead, removal commands use a weak degradation-cycle objective relative to the actual \(H_0\), addition commands use conservative class-aware synthetic targets, and the frozen grounding model verifies that each requested distortion changes by a sufficient margin without being replaced by another distortion. Appendix~\ref{app:osediff} gives the full definitions. Training starts from the public rank-4 OSEDiff checkpoint and uses Real-ESRGAN degradations, \(512\times512\) crops, effective batch size 40, Adam, a base learning rate of \(5\times10^{-5}\), 10 epochs, and 8 NVIDIA A100-80GB GPUs. The six new input columns use the base learning rate, whereas the remaining pretrained OSEDiff LoRA parameters use a reduced learning rate.

\begin{figure}[t]
\centering
\includegraphics[width=0.94\linewidth]{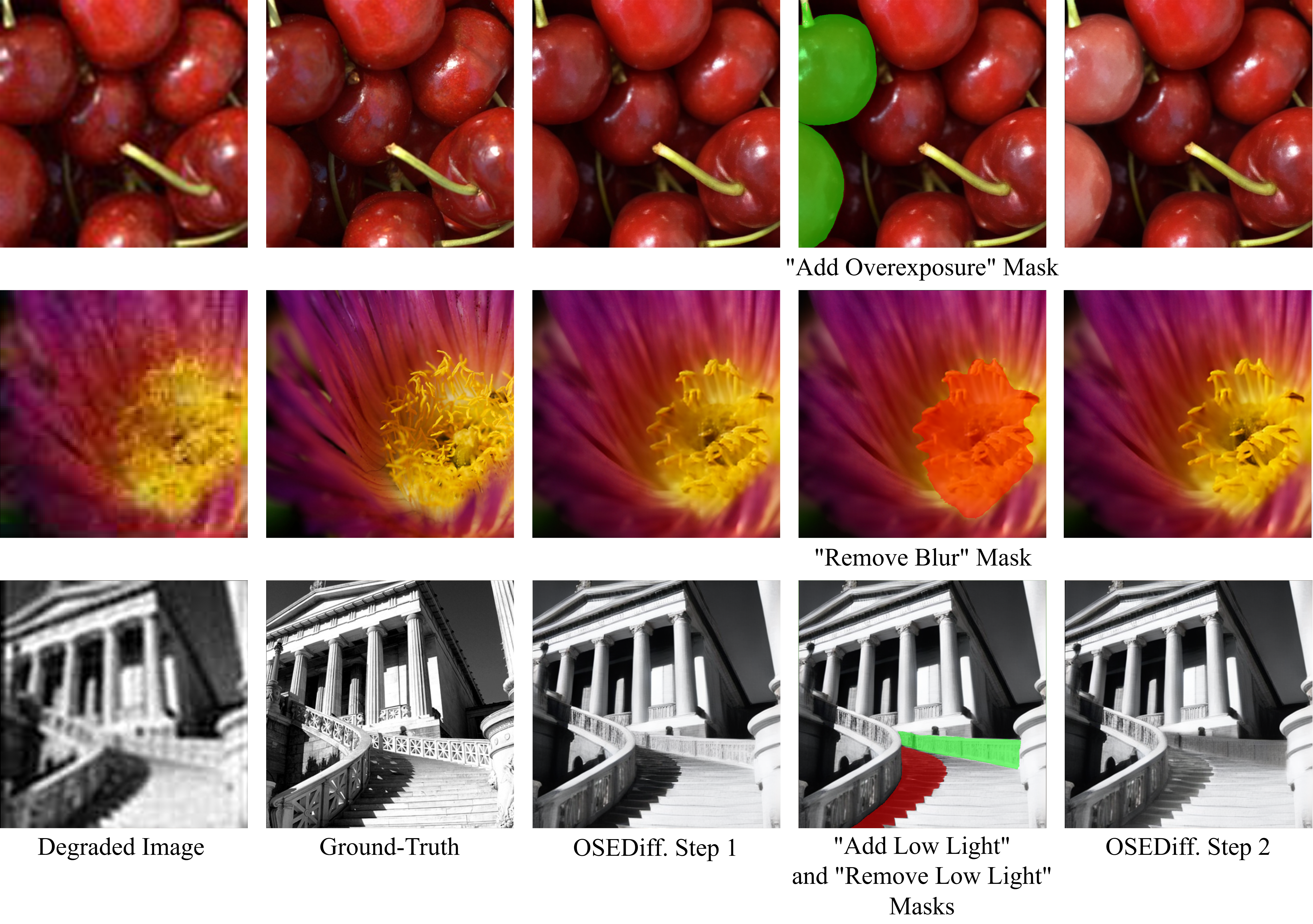}
\caption{Qualitative examples from the mask-conditioned OSEDiff application. They illustrate the spatial control interface and are not used as evidence for the SR-Ground benchmark results.}
\Description{Grid of low-resolution inputs and OSEDiff outputs under different spatial distortion-control masks.}
\label{fig:osediff-comparison}
\end{figure}

Figure~\ref{fig:osediff-comparison} presents qualitative outputs for several masks. The examples do not establish that every requested distortion can be reliably added or removed: severe information loss in the LR input can limit reconstruction, and errors from the frozen grounding model can propagate into the control signal.

\section{Conclusion}

In this work, we introduced SR-Ground, a dataset for image quality grounding tailored to super-resolution artifacts. We proposed an iterative curation pipeline that combines synthetic data generation, model-based annotation, and human-in-the-loop refinement via prominence scoring.

We found that SR-Ground improves SR-artifact localization in the evaluated settings and yields architecture-dependent transfer to Q-Ground. On a separate benchmark containing 1,000 outputs from five unseen SR methods, the RR model significantly improves prominence alignment over its NR counterpart and degrades when the LR reference is shuffled. Although Molodetskikh et al. remains strongest overall, the SR-Ground-trained RR model achieves the highest observed performance on native real-world datasets without HR ground truth. These results highlight both the value and the limits of domain-specific synthetic data for image quality grounding.

We believe SR-Ground provides a practical step toward more interpretable and fine-grained quality assessment, and can serve as a foundation for future research in super-resolution quality modeling.

%\begin{acks}
%This work was supported by the The Ministry of Economic Development of the Russian Federation in accordance with the subsidy agreement (agreement identifier 000000C313925P4H0002; grant No 139-15-2025-012).

%The research was carried out using the MSU-270 supercomputer of Lomonosov Moscow State University.
%\end{acks}

\bibliographystyle{ACM-Reference-Format}
\bibliography{samples/sample-base}

\clearpage
\appendix

\section{Dataset Curation Diagnostics}
\label{app:dataset-documentation}

\subsection{Responsible Use}

Crowdsourcing participants provided informed consent and received compensation through Yandex Tasks, and no worker personal data is stored. SR-Ground is intended for image restoration and quality assessment. Because mask-guided editing can also be used to alter visual evidence, applications should disclose such modifications.

\subsection{Filtering Q-Ground}

\begin{figure*}[t]
\centering
\includegraphics[width=0.32\textwidth]{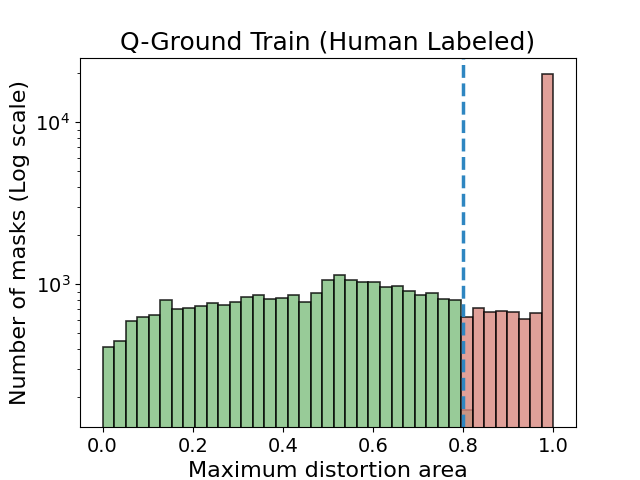}\hfill
\includegraphics[width=0.32\textwidth]{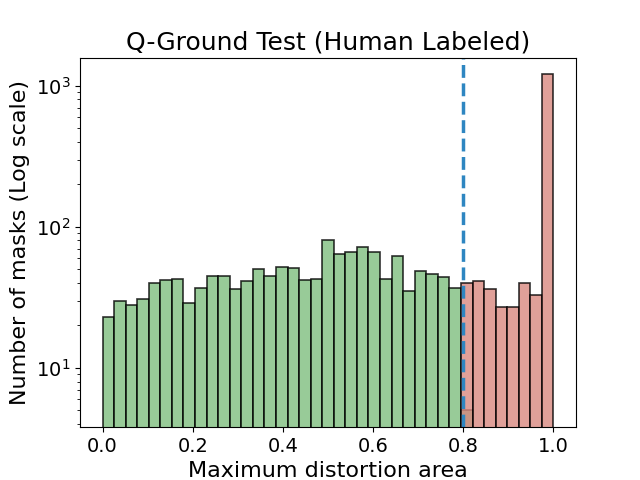}\hfill
\includegraphics[width=0.32\textwidth]{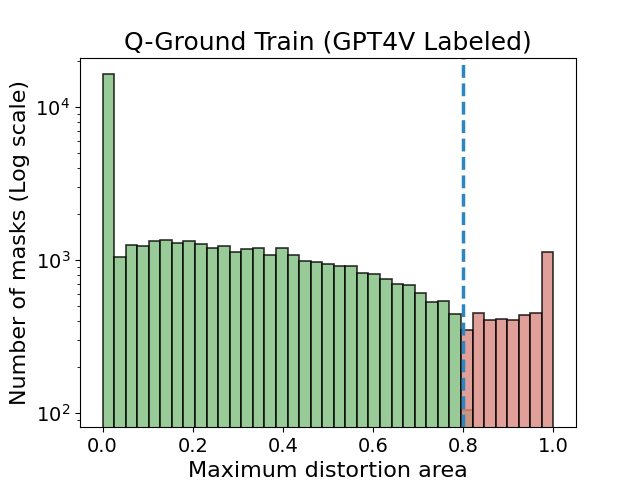}
\caption{Maximum distortion-area distributions for Q-Ground subsets. The dashed line marks the 0.8 filtering threshold.}
\label{fig:qground-filtering-histograms}
\end{figure*}

Q-Ground contains many annotations in which one distortion covers nearly the entire image. Such masks weaken locality and can encourage nearly constant predictions. We remove an annotation when its largest normalized class area exceeds 0.8; if an image has no annotations left, it is removed. Figure~\ref{fig:qground-filtering-histograms} shows the corresponding distributions for the human-labeled train and test subsets and the GPT4V-labeled train subset.

\subsection{Crowdsourcing Selection and Vote Budget}

\begin{figure*}[t]
\centering
\begin{minipage}[c]{0.67\textwidth}
  \centering
  \includegraphics[width=\linewidth]{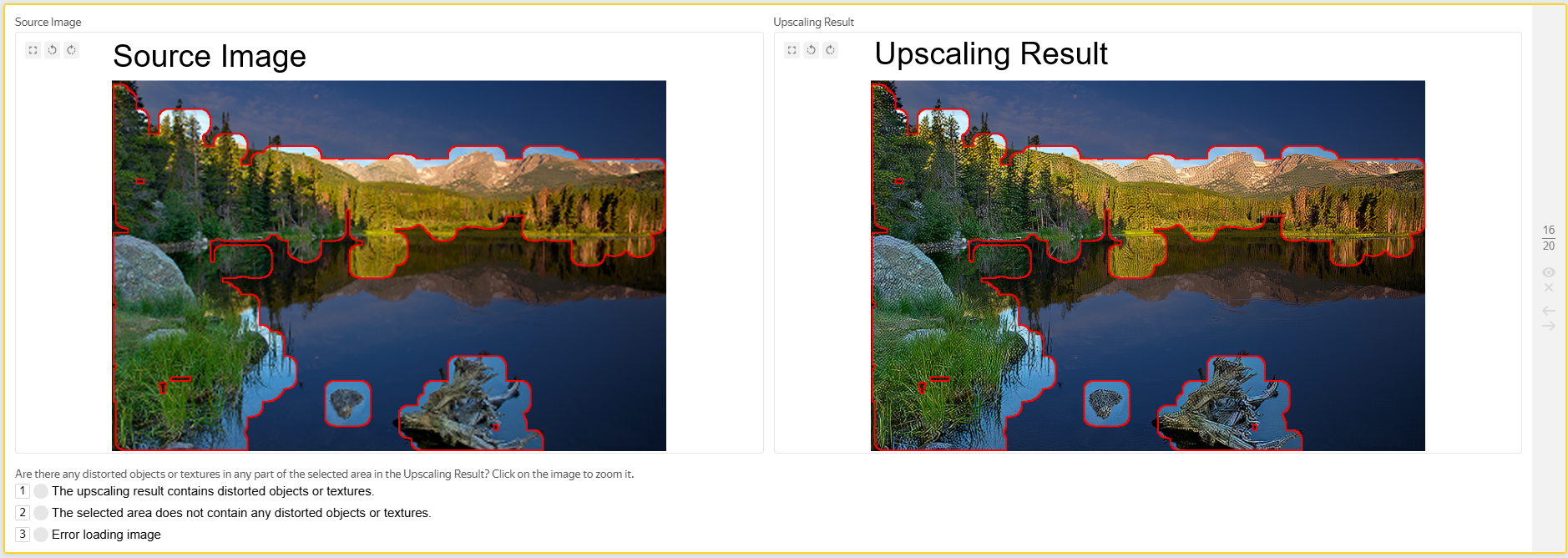}
\end{minipage}\hfill
\begin{minipage}[c]{0.30\textwidth}
  \centering
  \includegraphics[width=\linewidth]{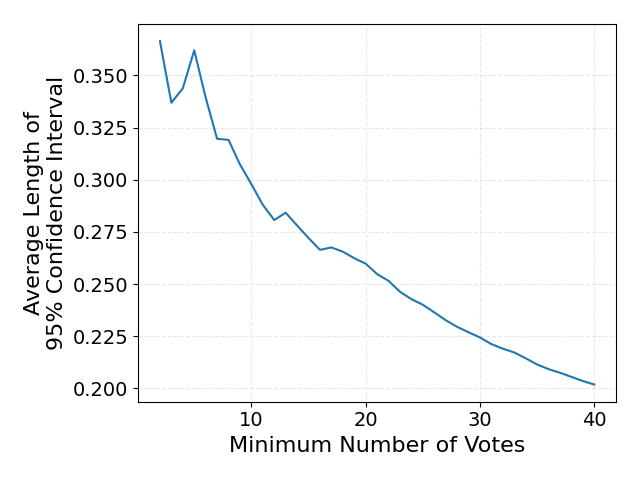}
\end{minipage}
\caption{Left: Yandex Tasks interface for deciding whether the highlighted region contains the named distortion. Right: average 95\% confidence-interval length for prominence versus the minimum number of votes.}
\Description{Crowdsourcing task interface beside a line plot showing how the average confidence-interval length decreases as the minimum vote count increases.}
\label{fig:crowdsourcing-interface-budget}
\end{figure*}

For a multi-class distortion output, we prioritize uncertain masks using
\[
\operatorname{score}(M,D,z)
=
\frac{1}{\log(1+|M_D|)}
\frac{1}{|M_D|}
\sum_{p\in M_D}
\left(\max\nolimits_1 z(p)-\max\nolimits_2 z(p)\right)
 .
\]
Here, \(M_D\) is the set of pixels predicted as class \(D\), and \(\max_1\) and \(\max_2\) are the two largest logits at a pixel. For the binary SR-artifact detector, whose binarization threshold is 0.3, we use
\[
\operatorname{score}(M,z)
=
\frac{1}{\log(1+|M_{\mathrm{SR}}|)}
\frac{1}{|M_{\mathrm{SR}}|}
\sum_{p\in M_{\mathrm{SR}}}
\left(z(p)-0.3\right)
 .
\]
We send the 10\% lowest-scoring blur and SR-artifact masks and the 5\% lowest-scoring low-light and overexposure masks to annotation, for 18,900 masks in total. A 40-vote pilot on approximately 1,000 samples per class determines the final budget of 10 votes per mask. Figure~\ref{fig:crowdsourcing-interface-budget} shows the annotation interface and the average 95\% confidence-interval length as a function of vote count.

The agreement values in Table~\ref{tab:annotation-agreement} are computed only for the four classes retained in crowdsourcing refinement. Because selection targets uncertain masks, agreement is lowest near prominence 0.5 and substantially higher for strong masks. This selection effect also explains why Krippendorff's \(\alpha\) is lower than raw pairwise agreement.

\begin{figure*}[t]
\centering
\includegraphics[width=0.98\textwidth]{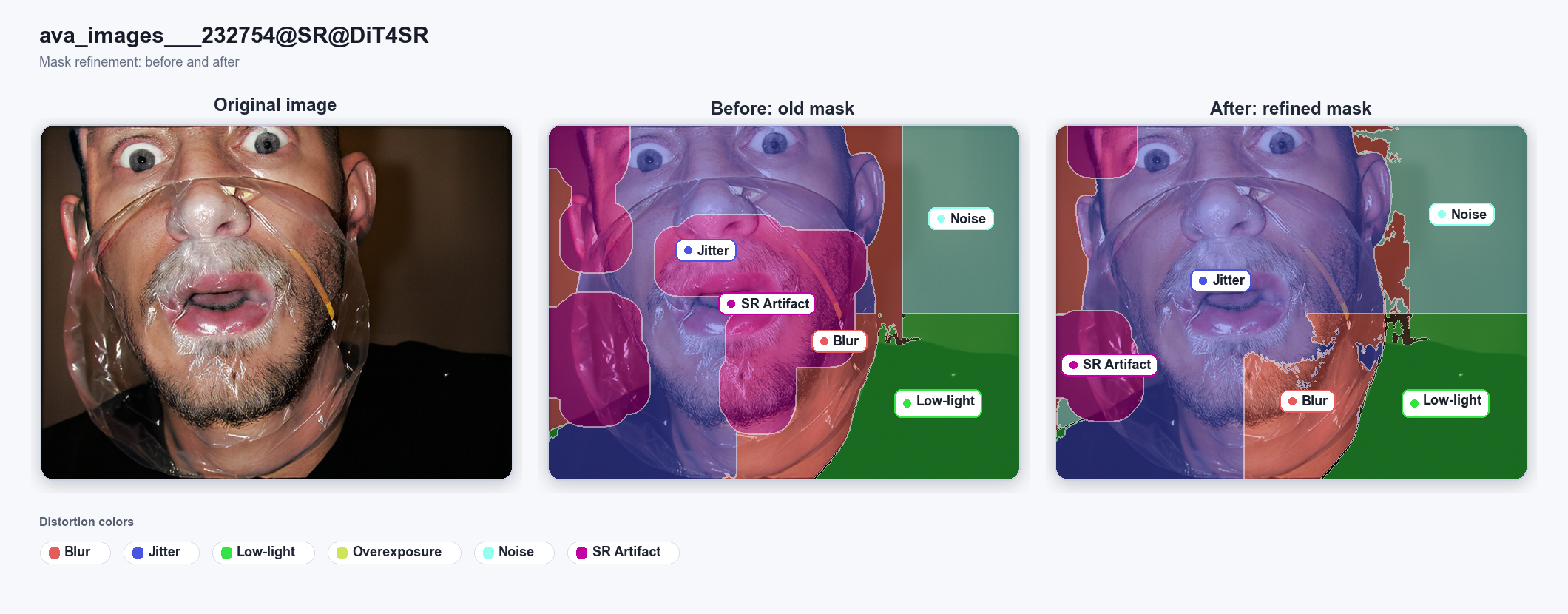}\\[2mm]
\includegraphics[width=0.98\textwidth]{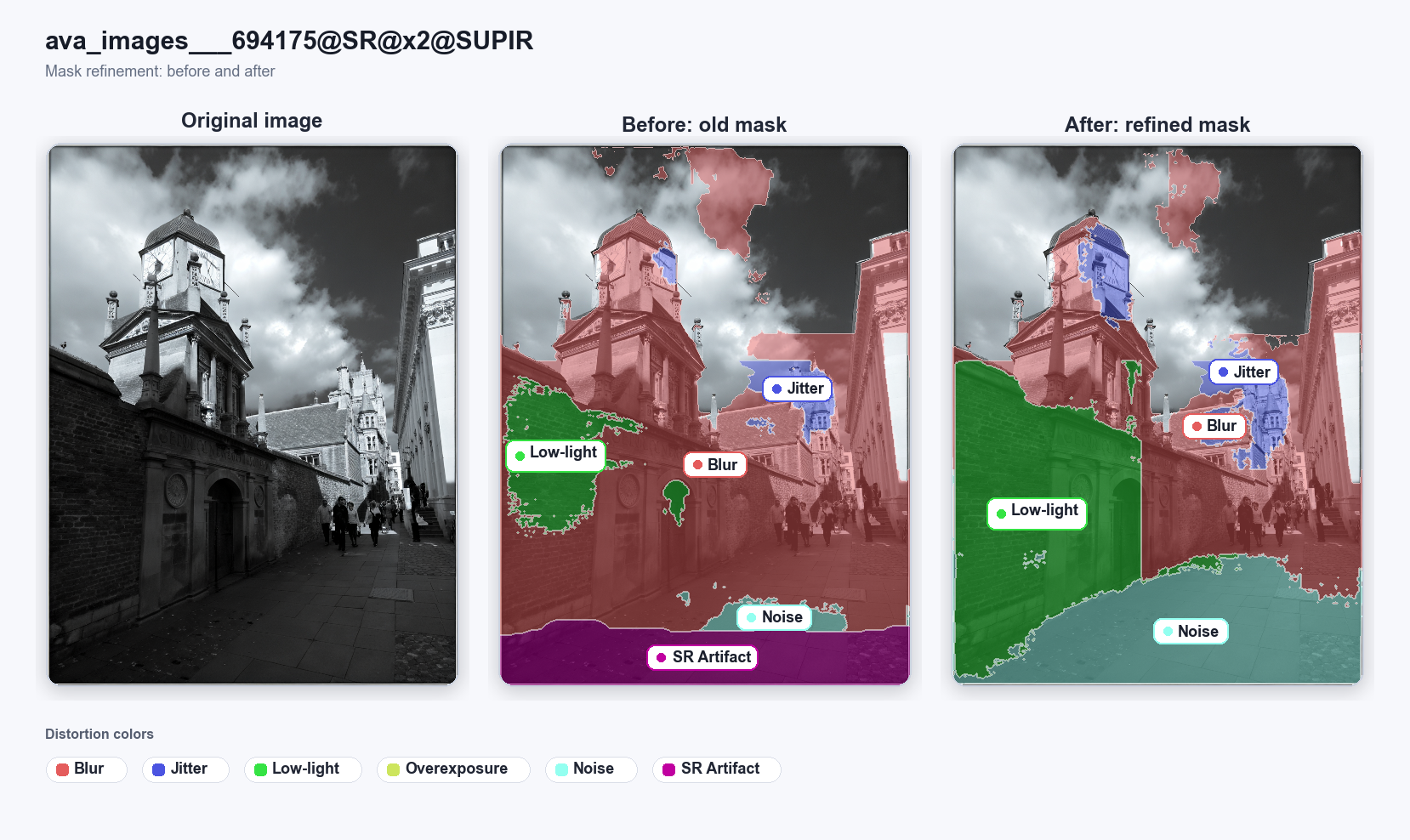}
\caption{Examples of model-generated masks before and after prominence-based refinement. Human feedback changes uncertain class assignments and boundaries; it does not replace every pixel with a manually drawn mask.}
\Description{Two examples comparing original model-generated distortion masks with masks after prominence-based refinement.}
\label{fig:refinement-examples}
\end{figure*}

\subsection{Artifact Statistics and IQA Correlation}

\begin{figure*}[t]
\centering
\includegraphics[width=0.92\textwidth]{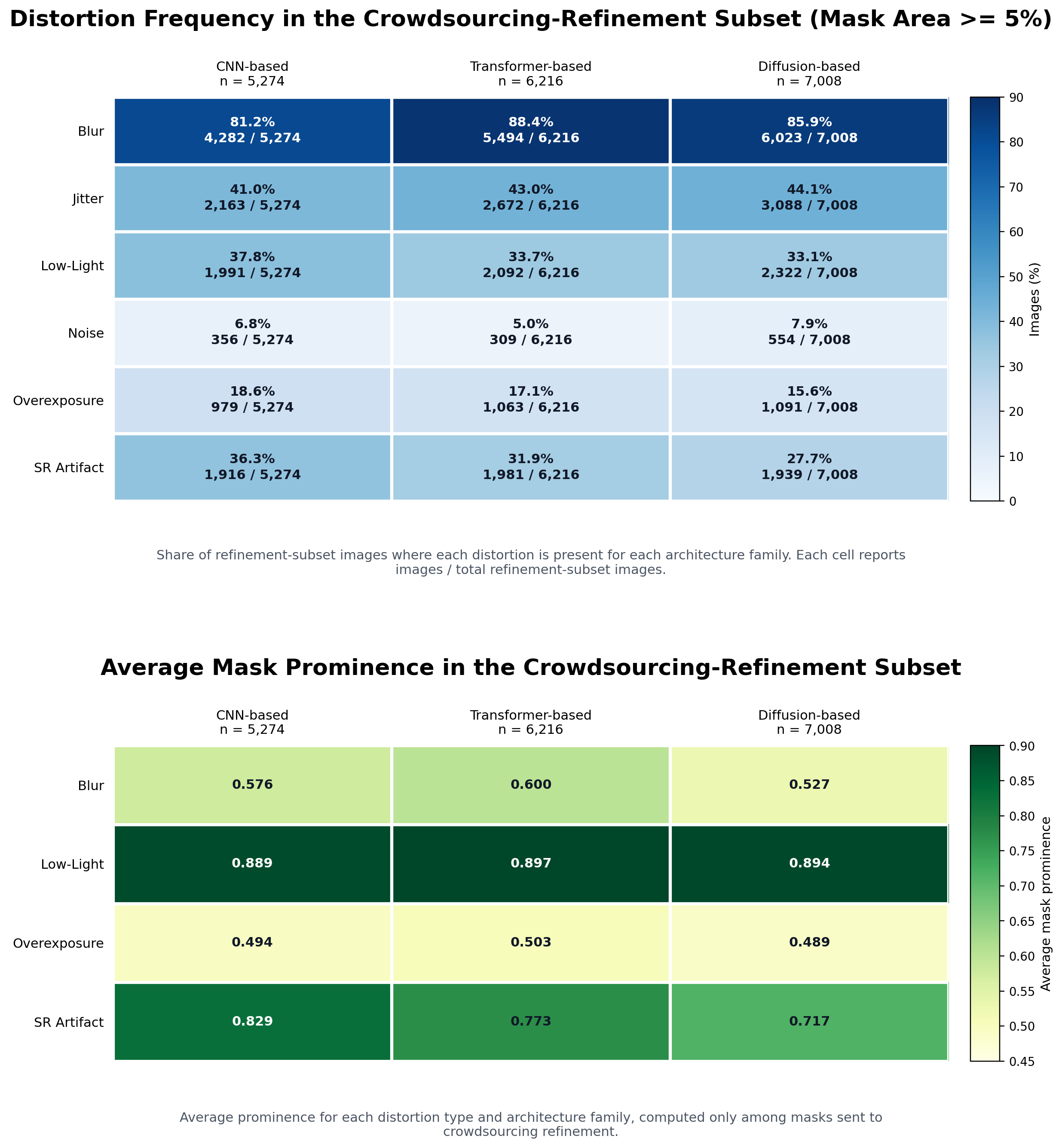}
\caption{Distortion frequency and mean mask prominence in the crowdsourcing-refinement subset, grouped by SR architecture family.}
\Description{Two heatmaps reporting distortion frequency and mean mask prominence across CNN-, transformer-, and diffusion-based super-resolution groups.}
\label{fig:artifact-prevalence-prominence}
\end{figure*}

Figure~\ref{fig:artifact-prevalence-prominence} reports frequency and average prominence in the crowdsourcing-refinement subset by SR architecture family. The statistics describe masks with area at least 5\% and should not be interpreted as the population frequency of every possible artifact. In this selected subset, blur is frequent across all families, while SR-artifact prominence is highest for the CNN-based group even though its frequency differs only moderately across families.

\begin{figure*}[t]
\centering
\includegraphics[width=0.98\textwidth]{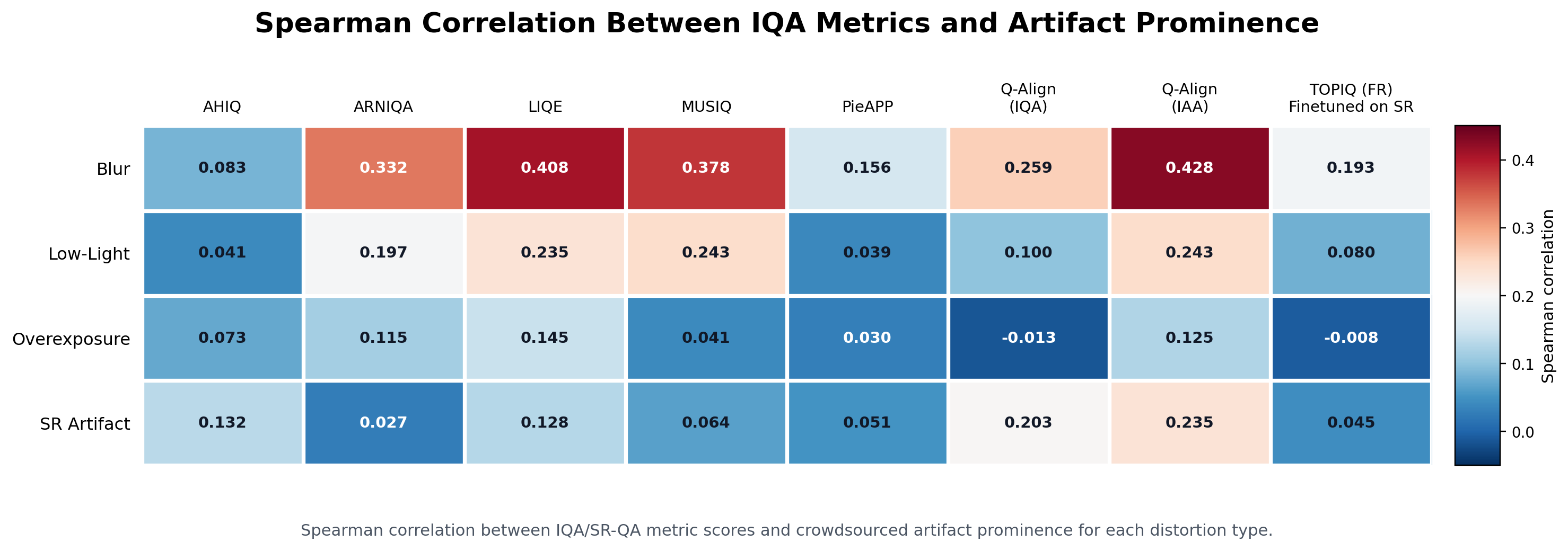}
\caption{Spearman correlation between holistic IQA/SR-QA scores and crowdsourced prominence for each refined distortion class.}
\Description{Heatmap of Spearman correlations between image-quality metrics and crowdsourced distortion prominence for blur, low light, overexposure, and SR artifacts.}
\label{fig:iqa-prominence-correlation}
\end{figure*}

We also compute Spearman correlation between crowdsourced prominence and eight holistic IQA/SR-QA scores. Figure~\ref{fig:iqa-prominence-correlation} shows weak or moderate correlations depending on the class and metric. The result motivates spatial, class-specific annotations, but it does not imply that holistic IQA is unsuitable for its intended global-quality task.

\subsection{Limitations}

The SR-artifact label remains heterogeneous because a stable pixel-level subtype taxonomy is unavailable. The dataset also covers a finite set of source collections, degradations, and SR systems, and some labels originate from models before refinement. Source-disjoint evaluation prevents direct image leakage but cannot establish robustness to every future SR architecture. Finally, jitter and noise do not receive crowdsourced prominence refinement because their pilot judgments were unreliable. These limitations should be considered when using SR-Ground as training data or as a benchmark.

\FloatBarrier
\section{Grounding-Model Training Details}
\label{app:grounding-training}

\subsection{Optimization Setup}

SegFormer and Mask2Former use batch size 40 and Adam with weight decay \(10^{-4}\). SegFormer uses learning rates \(10^{-5}\) for the encoder and \(5\times10^{-5}\) for the decoder; Mask2Former uses \(10^{-6}\) for the backbone and \(10^{-5}\) for the decoder. A cosine schedule includes a 10\% warmup over 50,000 iterations, although most runs converge within 10,000 iterations. Fine-tuning on non-SR data reduces every learning rate by a factor of 10. Training uses 8 NVIDIA A100-80GB GPUs.

Mask2Former starts from a Swin-Large backbone pretrained on Cityscapes. SegFormer-B5 starts from Cityscapes pretraining at \(1024\times1024\), with ADE20K pretraining used for \(448\times448\) runs. Augmentation is restricted to cropping, horizontal/vertical flipping, and 90-degree rotation. Binary cross-entropy plus Dice consistently outperforms cross-entropy plus Dice in the extended sweep. SegFormer performs best at \(448\times448\), while Mask2Former benefits from \(1024\times1024\).

\subsection{Extended Q-Ground Results}

\begin{table*}[t]
\centering
\caption{Extended Q-Ground results. U/F denote unfiltered/filtered training data; M/V denote manual-only/manual+GPT4V annotations. The best result is \textbf{bolded}, the second best is \underline{underlined} within groups of the same model type.}
\label{tab:extended-qground-results}
\scriptsize
\setlength{\tabcolsep}{3.5pt}
\begin{tabular}{l|l|c|c|c|c|c|c|c}
\hline
Model & Loss & Filter & Annot. & Resolution & Unf.\ mIoU$\uparrow$ & Unf.\ mAcc$\uparrow$ & Filt.\ mIoU$\uparrow$ & Filt.\ mAcc$\uparrow$ \\
\hline
SegFormer & BCE+Dice & U & M & 448 & \textbf{.573} & \textbf{.650} & .475 & .593 \\
SegFormer & CE+Dice & U & M & 448 & \underline{.567} & \underline{.648} & .473 & .596 \\
SegFormer & BCE+Dice & U & V & 448 & .547 & .620 & .446 & .559 \\
SegFormer & CE+Dice & U & V & 448 & .537 & .608 & .435 & .544 \\
SegFormer & BCE+Dice & F & M & 448 & .554 & .616 & \textbf{.527} & \underline{.619} \\
SegFormer & CE+Dice & F & M & 448 & .554 & .623 & \underline{.522} & \textbf{.623} \\
SegFormer & BCE+Dice & F & V & 448 & .507 & .564 & .472 & .555 \\
SegFormer & CE+Dice & F & V & 448 & .509 & .564 & .465 & .546 \\
SegFormer & BCE+Dice & U & M & 1024 & .530 & .625 & .411 & .557 \\
SegFormer & CE+Dice & U & M & 1024 & .526 & .620 & .405 & .550 \\
SegFormer & BCE+Dice & U & V & 1024 & .504 & .578 & .380 & .495 \\
SegFormer & CE+Dice & U & V & 1024 & .485 & .555 & .358 & .466 \\
SegFormer & BCE+Dice & F & M & 1024 & .528 & .594 & .486 & .585 \\
SegFormer & CE+Dice & F & M & 1024 & .524 & .588 & .473 & .570 \\
SegFormer & BCE+Dice & F & V & 1024 & .443 & .495 & .400 & .475 \\
SegFormer & CE+Dice & F & V & 1024 & .459 & .510 & .405 & .481 \\
\hline
Mask2Former & BCE+Dice & U & M & 448 & \underline{.586} & .657 & .496 & .604 \\
Mask2Former & CE+Dice & U & M & 448 & .553 & .625 & .454 & .563 \\
Mask2Former & BCE+Dice & U & V & 448 & .539 & .606 & .435 & .537 \\
Mask2Former & CE+Dice & U & V & 448 & .512 & .586 & .402 & .514 \\
Mask2Former & BCE+Dice & F & M & 448 & .559 & .619 & \underline{.530} & .619 \\
Mask2Former & CE+Dice & F & M & 448 & .523 & .576 & .486 & .563 \\
Mask2Former & BCE+Dice & F & V & 448 & .472 & .523 & .418 & .493 \\
Mask2Former & CE+Dice & F & V & 448 & .428 & .469 & .367 & .426 \\
Mask2Former & BCE+Dice & U & M & 1024 & \textbf{.591} & \textbf{.671} & .498 & \underline{.621} \\
Mask2Former & CE+Dice & U & M & 1024 & .580 & \underline{.665} & .480 & .611 \\
Mask2Former & BCE+Dice & U & V & 1024 & .539 & .609 & .426 & .534 \\
Mask2Former & CE+Dice & U & V & 1024 & .511 & .573 & .392 & .486 \\
Mask2Former & BCE+Dice & F & M & 1024 & .564 & .630 & \textbf{.534} & \textbf{.632} \\
Mask2Former & CE+Dice & F & M & 1024 & .535 & .593 & .496 & .583 \\
Mask2Former & BCE+Dice & F & V & 1024 & .496 & .549 & .463 & .540 \\
Mask2Former & CE+Dice & F & V & 1024 & .479 & .529 & .437 & .511 \\
\hline
\end{tabular}
\end{table*}

\begin{table*}[t]
\centering
\caption{Comparison with the Q-Ground paper on the unfiltered test set.}
\label{tab:qground-paper-comparison}
\scriptsize
\setlength{\tabcolsep}{3.5pt}
\begin{tabular}{l|c|c|c|c|c|c|c|c|c|c|c|c}
\hline
Method & \multicolumn{2}{|c|}{\cellcolor{jitter-color!75}jitter} & \multicolumn{2}{|c|}{\cellcolor{noise-color!75}noise} & \multicolumn{2}{|c|}{\cellcolor{overexposure-color!75}overexposure} & \multicolumn{2}{|c|}{\cellcolor{blur-color!75}blur} & \multicolumn{2}{|c|}{\cellcolor{lowlight-color!75}low light} & \multicolumn{2}{|c}{Average} \\\cline{2-13}
& mIoU$\uparrow$ & mAcc$\uparrow$ & mIoU$\uparrow$ & mAcc$\uparrow$ & mIoU$\uparrow$ & mAcc$\uparrow$ & mIoU$\uparrow$ & mAcc$\uparrow$ & mIoU$\uparrow$ & mAcc$\uparrow$ & mIoU$\uparrow$ & mAcc$\uparrow$ \\
\hline
SegFormer (Q-Ground) & .327 & .625 & .136 & .249 & .264 & .389 & .515 & 	.842 & .274 & .524 & .373 & .636 \\
Mask2Former (Q-Ground) & .401 & .625 & .089 & .113 & .223 & .424 & .566 & .902 & .290 & .461 & .403 & .646 \\
QG Model (Q-Ground) & .420 & .719 & .179 & .414 & .276 & .517 & .465 & .610 & .297 & .446 & .327 & .541 \\
\hline
SegFormer (ours) & .644 & .722 & .240 & .267 & .415 & .472 & .700 & .793 & .466 & .542 & .573 & .650 \\
Mask2Former (ours) & .663 & .732 & .271 & .317 & .494 & .563 & .709 & .811 & .436 & .497 & .591 & .671 \\
\hline
\end{tabular}
\end{table*}

The complete per-class table is available with the released evaluation files. Table~\ref{tab:qground-paper-comparison} retains the headline comparison while Table~\ref{tab:extended-qground-results} exposes the training choices needed to reproduce model selection.

\begin{table*}[t]
\centering
\caption{Ablation studies on train datasets, losses and used reference.}
\label{tab:ablations}
\tiny\begin{tabular}{l|c|c|c|c|c|c|c|c|c|c|c|c|c|c}
    \hline
    \multirow{2}{*}{Method} &
    \multirow{2}{*}{\makecell{Reference \\ Type}} &
    \multirow{2}{*}{Loss} &
    \multirow{2}{*}{Train Datasets} &
    \multirow{2}{*}{\makecell{SRCC \\ Open Images}$\uparrow$} &
    \multirow{2}{*}{\makecell{SRCC \\ Urban100}$\uparrow$} &
    \multirow{2}{*}{\makecell{SRCC \\ Urban100-HR}$\uparrow$} &
    \multirow{2}{*}{\makecell{SRCC \\ DeSRA}$\uparrow$} &
    \multirow{2}{*}{Avg.Rank$\downarrow$} &
    \multicolumn{4}{c|}{DeSRA} &
    \multicolumn{2}{c}{Q-Ground} \\
    \cline{10-15}
    & & & & & & & & & \makecell{F1 \\ (Full)}$\uparrow$ & \makecell{PR-AUC \\ (Full)}$\uparrow$ & \makecell{IoU \\ (Prominent)}$\uparrow$ & \makecell{PR-AUC \\ (Prominent)}$\uparrow$ & mIoU$\uparrow$ & mAcc$\uparrow$ \\
    \hline
    \makecell{No-Reference} & - & \makecell{BCE + \\ Dice + \\ KD} & \makecell{Q-Ground + \\ Open Images} & 0.359 & 0.274 & 0.120 & 0.201 & 6.00 & 0.157 & 0.035 & 0.392 & 0.305 & 0.516 & 0.603 \\
    \hline
    \makecell{No-Reference} & - & \makecell{BCE + \\ Dice +  \\ KD} & \makecell{Q-Ground + \\ Open Images + \\ SR-Ground} & 0.353 & 0.178 & 0.185 & 0.211 & 6.25 & 0.161 & 0.034 & 0.388 & 0.335 & \underline{0.526} & \underline{0.620} \\
    \hline
    \makecell{Backbone Fusion} & Bicubic & \makecell{BCE + \\ Dice +  \\ KD + \\ Identity} & \makecell{Q-Ground + \\ Open Images + \\ SR-Ground} & 0.291 & 0.362 & 0.246 & 0.255 & 5.00 & 0.182 & 0.042 & 0.452 & 0.401 & \textbf{0.529} & \textbf{0.625} \\
    \hline
    \makecell{Backbone Fusion} & RLFN & \makecell{BCE + \\ Dice} & \makecell{Q-Ground + \\ Open Images + \\ SR-Ground} & 0.356 & \textbf{0.416} & \textbf{0.412} & \underline{0.292} & \underline{2.25} & 0.189 & 0.048 & 0.473 & 0.457 & 0.490 & 0.560 \\
    \hline
    \makecell{Backbone Fusion} & RLFN & \makecell{BCE + \\ Dice +  \\ KD} & \makecell{Q-Ground + \\ Open Images + \\ SR-Ground} & \underline{0.386} & \underline{0.406} & 0.381 & 0.281 & 2.50 & 0.190 & \underline{0.052} & \underline{0.486} & \underline{0.494} & 0.493 & 0.566 \\
    \hline
    \makecell{Backbone Fusion} & RLFN & \makecell{BCE + \\ Dice +  \\ KD + \\ Identity} & \makecell{Q-Ground + \\ Open Images} & 0.371 & 0.402 & 0.304 & \textbf{0.311} & 3.00 & \underline{0.191} & \textbf{0.059} & 0.400 & 0.467 & 0.501 & 0.576 \\
    \hline
    \makecell{Backbone Fusion} & RLFN & \makecell{BCE + \\ Dice +  \\ KD + \\ Identity} & \makecell{Q-Ground + \\ Open Images + \\ SR-Ground} & \textbf{0.395} & \underline{0.406} & \underline{0.397} & \textbf{0.311} & \textbf{1.50} & \textbf{0.193} & 0.050 & \textbf{0.489} & \textbf{0.627} & 0.512 & 0.597 \\
    \hline
\end{tabular}
\end{table*}

\subsection{Ablation Studies}
Table~\ref{tab:ablations} shows ablation studies for No-Reference model and Reduced-Reference Backbone Fusion model. Training on Open Images dataset only, without Identity Loss, without boss Identity Loss and KD Loss, with reference obtained with bicubic upscaling were tested.
 
\section{OSEDiff Losses and Training Details}
\label{app:osediff}

\begin{figure*}[t]
\centering
\includegraphics[width=0.75\linewidth]{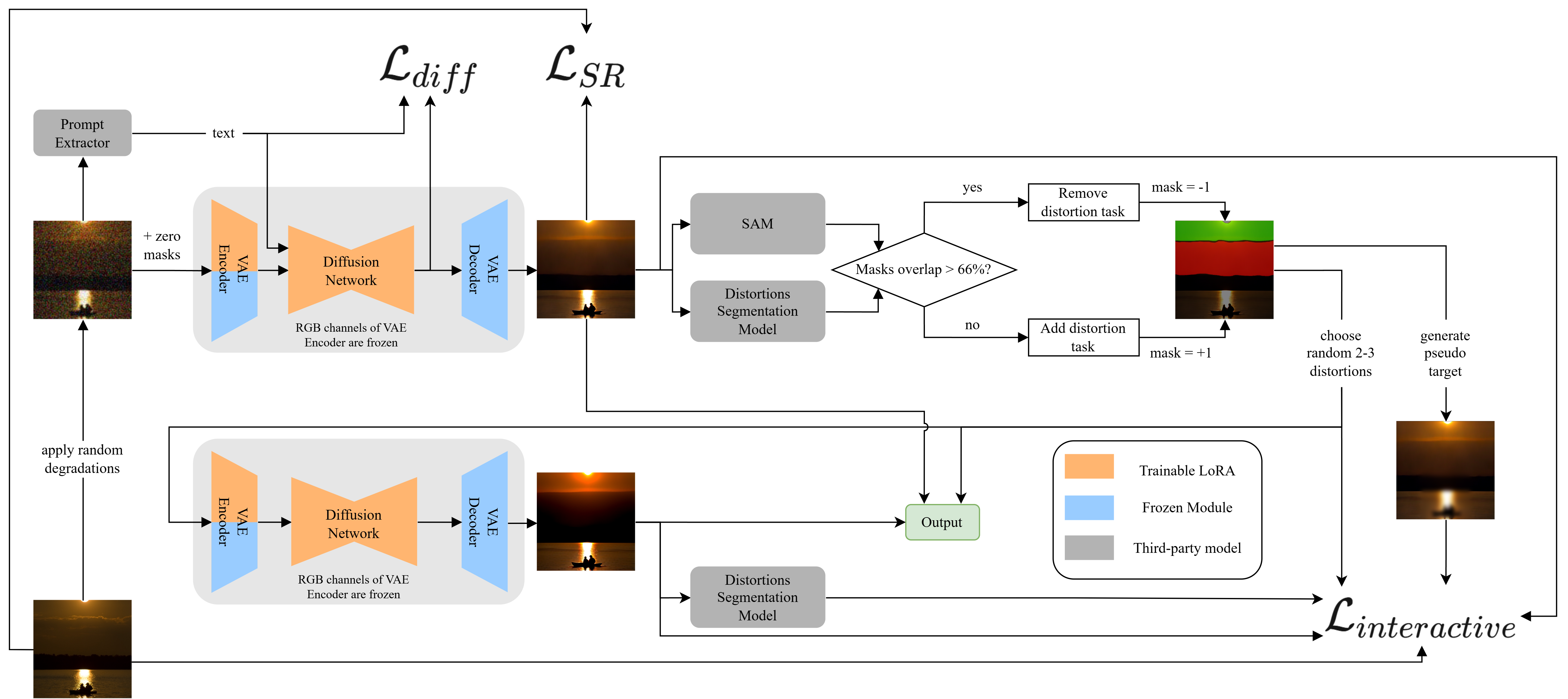}
\caption{Mask-conditioned two-pass OSEDiff training pipeline.}
\Description{Two-pass OSEDiff pipeline in which the first pass performs standard super-resolution and the second pass applies spatially restricted signed distortion commands verified by a frozen grounding model.}
\label{fig:osediff-training}
\end{figure*}

Section~\ref{sec:osediff-application} presents the pipeline, its training principles, training configuration, and qualitative examples. This appendix specifies the conditioning architecture, mask-sampling procedure, and complete objective. Full training pipeline is depicted on Figure~\ref{fig:osediff-training}.

\subsection{\texorpdfstring{Mask Conditioning and Two-Pass Training}{Mask Conditioning and Two-Pass Training}}

We augment the Stable Diffusion 2.1 VAE encoder in OSEDiff~\cite{wu2024one,Rombach_2022_CVPR} with six channels ordered as blur, jitter, low-light, noise, overexposure, and SR artifact. If \(W_{\mathrm{RGB}}\) denotes the pretrained input-convolution weights, the expanded convolution is initialized as
\[
W_{\mathrm{in}}=[W_{\mathrm{RGB}},0_C].
\]
Only the six new full-rank columns are directly optimized, adding 6,912 trainable parameters. The RGB columns and the LoRA-A/LoRA-B parameters of this input convolution are frozen, while the remaining pretrained OSEDiff LoRA parameters continue training with a reduced learning rate. We do not create a new LoRA on top of the pretrained OSEDiff LoRA. Consequently, zero-valued command channels do not change the original model behavior.

The first pass produces
\[
H_0=G_\theta(x_{\mathrm{LQ}},0).
\]
A frozen grounding model predicts distortion maps, and SAM extracts at most 30 candidate regions occupying between 1\% and 80\% of the image. A region whose overlap with predicted distortion class \(k\) exceeds 66\% receives a removal request for that class; otherwise, it receives an addition request for a class that is absent or weakly represented in the region. We sample two or three distinct commands, reject strongly overlapping masks, and limit their total area. In the implementation, selected masks have pairwise IoU at most 0.35 and their union covers at most 75\% of the image. If any GPU cannot construct at least two valid commands, the corresponding distributed batch is used only for first-pass SR training. Commands remain separate during verification even when several masks are assigned to the same class.

Let \(C^{\mathrm{hard}}\in\{-1,0,+1\}^{B\times6\times H\times W}\) contain the binary SAM commands. These hard masks are used for class assignment and regional verification. The generator receives a soft map \(C\): values inside each selected region remain \(-1\) or \(+1\), while their magnitude decreases linearly to zero outside the region. For \(512\times512\) training crops, the feathering radius is 10 pixels, giving magnitude 0.5 at a distance of 5 pixels and zero at 10 pixels. If a low-light or overexposure region covers at least 25\% of the image, a weak global context of magnitude 0.15 is added with the same sign, while full command strength remains restricted to the hard region.

The second pass receives the actual first result, without synthetic pre-degradation:
\[
X_1=H_0,\qquad R_1=G_\theta(X_1,C).
\]
The raw output is spatially gated before verification and before being returned at inference:
\[
S=\max_k|C_k|,\qquad H_1=X_1+S\odot(R_1-X_1).
\]
Thus, \(H_1=X_1\) exactly outside the soft command support, the full edit is used in the hard region, and its strength decreases smoothly in the feathered ring. An explicitly supplied zero map on the second pass therefore requests identity, whereas the first pass without a supplied command map still performs normal super-resolution.

\subsection{\texorpdfstring{Local Edit Supervision}{Local Edit Supervision}}

The first pass retains the original OSEDiff supervision with full strength. Using the weights from our training setup, its image-space objective is
\[
\mathcal{L}_{\mathrm{SR}}
=
2\mathcal{L}_{\mathrm{L1}}(H_0,x_{\mathrm{GT}})
+3\mathcal{L}_{\mathrm{LPIPS}}(H_0,x_{\mathrm{GT}})
+\mathcal{L}_{\mathrm{reg}}(H_0),
\]
while the OSEDiff regulator diffusion loss \(\mathcal{L}_{\mathrm{diff}}\) is also computed only from the first-pass latent. The second pass has no \(H_1\!\rightarrow x_{\mathrm{GT}}\) loss, because the dataset ground truth is not assumed to be an upper bound on quality, and no mask-blind VSD is applied to \(H_1\). The raw output \(R_1\) is additionally constrained to reproduce \(X_1\) outside the entire soft command support.

For a removal command of class \(k\), let \(A_j\) be its hard SAM region and let \(D_k^-\) be a weak, near-identity forward distortion. The edited result is required to be one inverse step cleaner than the actual first pass:
\[
\mathcal{L}_{\mathrm{remove},j}
=
\mathcal{L}_{\mathrm{Charb}}
\!\left(D_k^-(H_1),\operatorname{sg}(H_0);A_j\right)
+0.1\,\mathcal{L}_{\mathrm{L1}}
\!\left(H_1,\operatorname{sg}(H_0);\widetilde{A}^{-}\right),
\]
where $\mathcal{L}_{\mathrm{Charb}}$ means the Charbonnier loss, \(\operatorname{sg}\) stops the gradient, losses are normalized by the selected area, and \(\widetilde{A}^{-}\) is the full soft removal support after excluding pixels occupied by an addition command. The first term is evaluated in the hard core; for spatial convolutional operators, the core is eroded by two pixels to avoid a training contour at the boundary. The second term selects the minimum sufficient change and limits ringing or overshoot.

The conservative forward operators are residual Gaussian blur with coefficient 0.20 for blur, symmetric one-pixel chromatic ghosting with coefficient 0.12 for jitter, the reversible curve \(x^{1.15}\) for low-light, the reversible curve \(1-(1-x)^{1.15}\) for overexposure, and an unsharp-forward operator with coefficient 0.25 for SR artifacts. An additive-noise cycle is not used because its expected optimum is identity. Instead, removal of noise uses the pseudo-target
\[
0.65H_0+0.35\operatorname{median}_{3\times3}(H_0),
\]
where $\operatorname{median}_{3\times3}$ denotes median filtering with a $3\times3$ kernel.

The removal objective is computed separately for every selected command before averaging.

For an addition command, a conservative class-aware target \(D_k(H_0)\) is created: Gaussian blur for blur, small ghosting and chromatic misregistration for jitter, darkening with moderate shot/read noise for low-light, controlled noise for noise, increased exposure with clipping and weak bloom for overexposure, and moderate oversharpening/ringing for SR artifacts. The raw output \(R_1\), rather than the blended \(H_1\), is trained to reproduce the full local transformation; residual gating subsequently sets its spatial strength.

For noise addition, exact pixel-wise regression would average toward zero. We therefore supervise the mean and variance of the residual, its low-frequency energy, and short-range correlations. A fresh random pixel target is added with weight 0.05 only to escape the stationary zero-residual solution; the sampled realization changes for every example, while the correlation term discourages a deterministic or repeating pattern. These synthetic targets provide a physically meaningful edit direction but do not replace grounding-model verification.

\subsection{\texorpdfstring{Grounding-Based Verification}{Grounding-Based Verification}}

The Mask2Former model trained for image quality grounding predicts probabilities for seven classes at every input pixel: background and the six distortions. The baseline map is always evaluated on the actual \(H_0\) and detached from the computation graph. Let \(P_k(I)\) denote the probability of distortion \(k\), and define
\[
q_k(I)=P_k(I)-P_{\mathrm{bg}}(I).
\]
For addition and removal commands, respectively, the directed changes are
\[
\delta_{\mathrm{add},k}
=P_k(H_1)-\operatorname{sg}(P_k(H_0)),
\qquad
\delta_{\mathrm{remove},k}
=\operatorname{sg}(q_k(H_0))-q_k(H_1).
\]

A command is considered satisfied only when its directed change reaches a margin. For every command region \(A_j\), we use the area-normalized smooth-margin loss
\[
\mathcal{L}_{\mathrm{dir},j}
=
\frac{1}{|A_j|}
\sum_{p\in A_j}
\frac{\operatorname{softplus}\!\left(\beta(m_s-\delta_j(p))\right)}{\beta},
\]
with \(\beta=8\), \(m_{\mathrm{add}}=0.15\), and \(m_{\mathrm{remove}}=0.08\). Each SAM command is normalized and evaluated separately, so a successful command cannot compensate for an unsuccessful one.

For an addition command, the selected class is also ranked above all other classes with margin 0.05. For removal, background is ranked only against the requested class with the same margin; this term has weight 0.25 and does not require unrelated distortions to decrease. A second one-sided term, also weighted by 0.25, penalizes increases of any of the other five distortion probabilities beyond 0.02 relative to \(H_0\). Consequently, replacing blur with an SR artifact does not satisfy a blur-removal command. The image quality grounding model remains frozen, but gradients through its prediction on \(H_1\) update the generator.

The verifier weight is increased gradually during training and capped according to the local teacher losses. If a batch contains a removal command, its effective weight is lower-bounded by 0.02 so that the verifier gradient does not vanish near the identity solution \(H_1=H_0\). The complete second-pass objective is
\[
\mathcal{L}_{\mathrm{interactive}}
=
\mathcal{L}_{\mathrm{preserve}}
+\mathcal{L}_{\mathrm{remove}}
+\mathcal{L}_{\mathrm{add}}
+\lambda_{\mathrm{ver}}(t)\mathcal{L}_{\mathrm{ver}}.
\]
The same RAM~\cite{zhang2023recognize} content caption is retained for both passes. The second-pass negative prompt is neutral and does not prohibit distortion classes that may be requested by an addition command.

\end{document}